%% file: neurips_2026.tex
\documentclass{article}

\usepackage[main,final]{neurips_2026}

\usepackage{microtype}
\usepackage{graphicx}
\usepackage{subcaption}
\usepackage{booktabs} % for professional tables
\usepackage{comment}

\usepackage[utf8]{inputenc} % allow utf-8 input
\usepackage[T1]{fontenc}    % use 8-bit T1 fonts
\usepackage{hyperref}       % hyperlinks
\usepackage{url}            % simple URL typesetting
\usepackage{booktabs}       % professional-quality tables
\usepackage{amsfonts}       % blackboard math symbols
\usepackage{nicefrac}       % compact symbols for 1/2, etc.
\usepackage{microtype}      % microtypography
\usepackage{xcolor}         % colors

\usepackage{amsmath}
\usepackage{amssymb}
\usepackage{mathtools}
\usepackage{amsthm}
\usepackage{multicol}
\usepackage{multirow}
\usepackage{pifont} 

\usepackage{subcaption}

\theoremstyle{plain}

\theoremstyle{definition}

\theoremstyle{remark}

\usepackage[dvipsnames]{xcolor}

\usepackage{float}

\usepackage[capitalize,noabbrev]{cleveref}

\usepackage[dvipsnames]{xcolor}
\hypersetup{
  colorlinks,
  linkcolor={red!100!black},
  citecolor={blue!140!black},
  urlcolor={red!100!black}
  }
  \definecolor{orange}{HTML}{ff7f0e}
  \definecolor{blue}{HTML}{1f77b4}

\usepackage{enumitem}

\usepackage[table]{xcolor}
\definecolor{color3}{rgb}{0.92,0.92,0.92}

\definecolor{color2}{rgb}{0.62,0.45,0.45}

\usepackage{wrapfig}
\usepackage{titlesec}

\titlespacing*{\subsubsection}
{0pt}{0.7ex plus 0.2ex minus 0.1ex}{0.35ex plus 0.1ex}

\AtBeginDocument{
  \setlength{\abovedisplayskip}{4pt}
  \setlength{\belowdisplayskip}{4pt}
  \setlength{\abovedisplayshortskip}{2pt}
  \setlength{\belowdisplayshortskip}{2pt}
}

\makeatletter
\@ifundefined{iflatexml}{\newif\iflatexml}{}
\makeatother

\title{CoScan: Multi-Scale Content-Adaptive Space-Filling Scans for Causal State-Space Image Restoration}

\iflatexml
\author{%
Wei Dong\\
McMaster University\\
\texttt{dongw22@mcmaster.ca}\\
Equal Contribution
\And
Terry Ji\\
McMaster University\\
\texttt{jit7@mcmaster.ca}\\
Equal Contribution
\And
Yan Min\\
McMaster University\\
\texttt{miny13@mcmaster.ca}
\And
Shahram Shirani\\
McMaster University\\
\texttt{shirani@mcmaster.ca}
\And
Jun Chen\\
McMaster University\\
\texttt{chenjun@mcmaster.ca}
\And
Han Zhou\\
McMaster University\\
\texttt{zhouh115@mcmaster.ca}\\
Corresponding Author, Project Lead
}
\else
\author{%
  Wei Dong$^{*}$, Terry Ji$^{*}$, Yan Min, Shahram Shirani, Jun Chen, Han Zhou$^{\dagger}$ \\
  McMaster University \\
  {\{dongw22, jit7, miny13, shirani, chenjun, zhouh115\}}@mcmaster.ca \\
  $^*$Equal Contribution \quad $^\dagger$Corresponding Author, Project Lead
}
\fi

\usepackage{subcaption}

\begin{document}

\maketitle

 {\centering

\captionsetup{type=figure}
\captionsetup[subfigure]{skip=1pt}
\setcounter{figure}{0}
\renewcommand{\arraystretch}{0.95}

\begin{tabular}{@{}c@{\hspace{1.2mm}}c@{}}

\begin{minipage}[t]{0.48\textwidth}
    \vspace{0pt}
    \centering
\includegraphics[width=\linewidth,height=0.18\textheight,keepaspectratio]{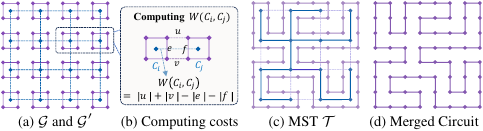}
    \subcaption*{\scriptsize \textbf{I. Cover-and-merge space-filling curve (SFC) construction}. The image is represented as a grid graph $\mathcal{G}$, with pixels covered by local $2\times2$ circuits $\{C_i\}$ (purple). The dual graph $\mathcal{G}^{\prime}$ (blue) connects adjacent circuits, and a minimum spanning tree (MST) on $\mathcal{G}^{\prime}$ induces a Hamiltonian circuit~\cite{nafnet,wang2022neural}.
    }
\end{minipage}
&
\begin{minipage}[t]{0.48\textwidth}
    \vspace{0pt}
    \centering
    \includegraphics[width=\linewidth,height=0.18\textheight,keepaspectratio]{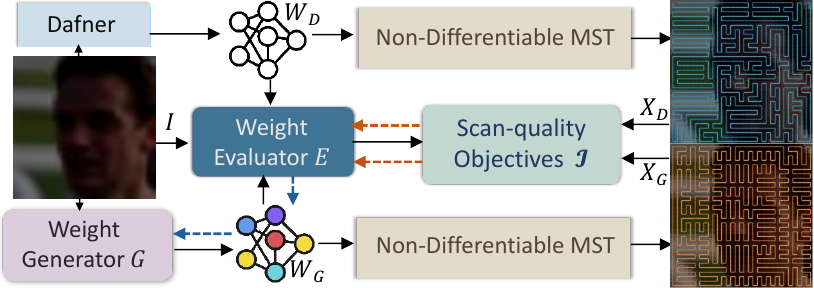}   
%    \vspace{-1.5mm}
    \subcaption*{\scriptsize \textbf{II. Coherence-preserving optimization of our content-aware SFCs}. The red and blue dashed arrows represent the gradient flow when optimizing $E$ or $G$, respectively.}
    
\end{minipage}
\\[-1.5mm]

\begin{minipage}[t]{0.48\textwidth}
    \vspace{0pt}
    \centering
    \includegraphics[width=\linewidth,height=0.17\textheight,keepaspectratio]{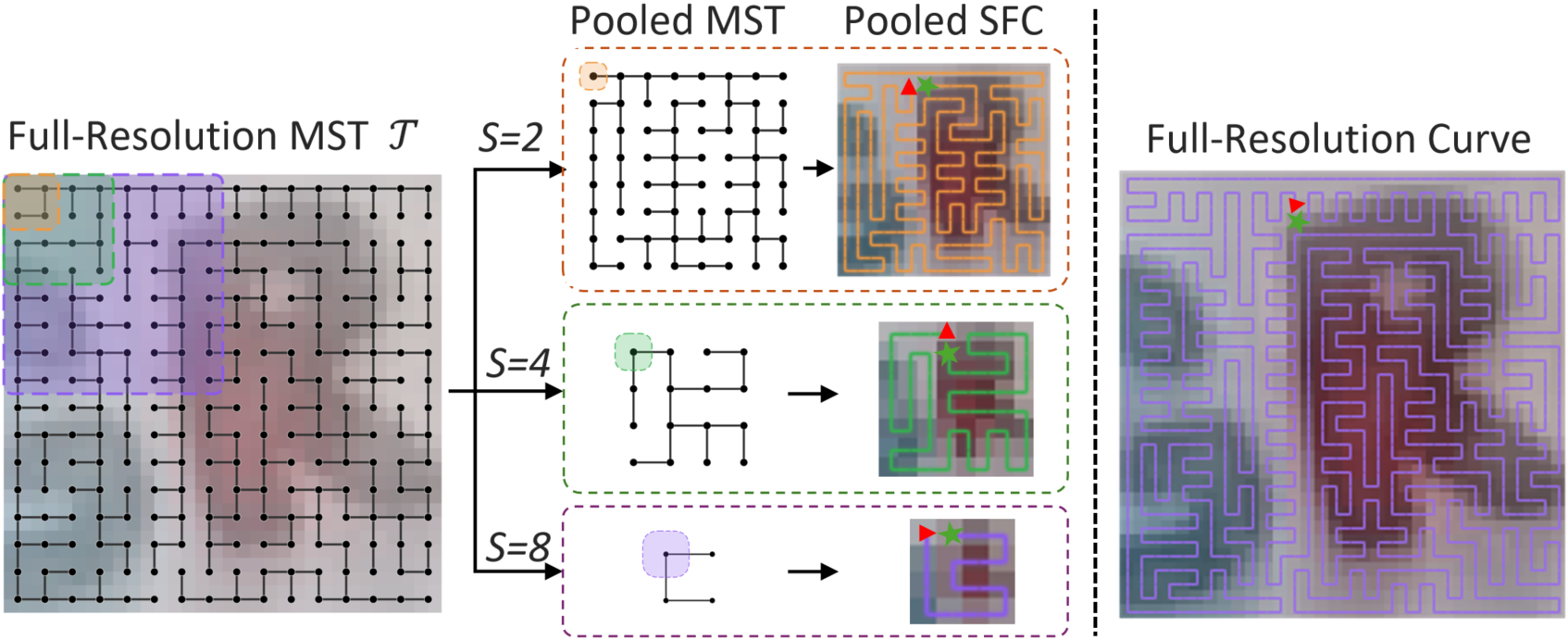}
    
%    \vspace{-1mm}
    \subcaption*{\scriptsize \textbf{III. Multi-scale scan scaling}. Our graph-based MST coarsening derives structure-consistent scans across feature resolutions.}
\end{minipage}
&
\begin{minipage}[t]{0.47\textwidth}
    \vspace{0pt}
    \centering
    \includegraphics[width=\linewidth,height=0.17\textheight,keepaspectratio]{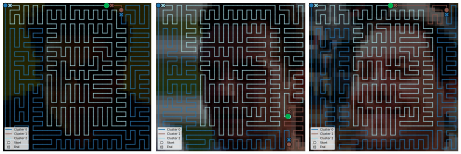}    
%    \vspace{-1mm}
    \subcaption*{\scriptsize \textbf{IV. Start selection on optimized SFCs}. High-frequency responses along the optimized SFCs from II are clustered into segments, and the segment with the largest frequency-weighted length score determines the curve start.  {\color{ForestGreen}\small$\bullet$} denotes the final selected curve start.}
\end{minipage}

\end{tabular}

\vspace{-1mm}
\addtocounter{figure}{-1}
\captionof{figure}{Overview of \textbf{CoScan}. Starting from a cover-and-merge construction for space-filling scans (\textbf{I}), CoScan extends it with differentiable content-aware optimization to obtain coherent and locality-preserving scan orders (\textbf{II}). The optimized scans are further adapted across feature resolutions (\textbf{III}), with informative starting points selected to strengthen causal vision Mamba modeling (\textbf{IV}).
}
\label{fig_teaser}

}
\vspace{4mm}
\input{sec/0_abstract}

\begin{figure}[t]
    \centering
    \setlength{\abovecaptionskip}{0.05in}
    \centering\setlength{\baselineskip}{0mm}
    \includegraphics[width=1\textwidth]{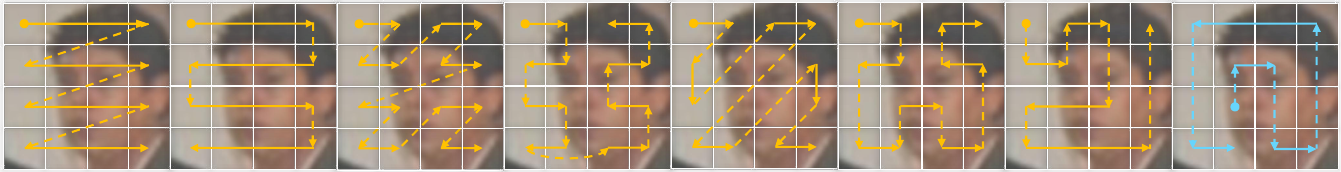}\par
   
    \begin{minipage}{1\textwidth}\centering

    \begin{minipage}[t]{0.123\linewidth}\centering
   
    \scriptsize (a) VMamba\\~\citep{liu2024vmamba}\end{minipage}%
\begin{minipage}[t]{0.123\linewidth}\centering\scriptsize (b) Zigma\\~\citep{hu2024zigma}\end{minipage}%
\begin{minipage}[t]{0.123\linewidth}\centering\scriptsize (c) LocalMamba\\~\citep{huang2024localmamba}\end{minipage}%
\begin{minipage}[t]{0.123\linewidth}\centering\scriptsize (d) MaIR\\~\citep{li2025mair}\end{minipage}%
\begin{minipage}[t]{0.123\linewidth}\centering\scriptsize (e) Diagonal\\~\citep{li2025osmamba}\end{minipage}%
\begin{minipage}[t]{0.123\linewidth}\centering\scriptsize (f) Hilbert\\(AAAI25~\citep{xiao2025boosting})\end{minipage}%
\begin{minipage}[t]{0.123\linewidth}\centering\scriptsize (g) NeuralSFC\\~\citep{wang2022neural}\end{minipage}%
\begin{minipage}[t]{0.123\linewidth}\centering\scriptsize (h) \textbf{CoScan}\\\textbf{(Ours)}\end{minipage}

\end{minipage}
\vspace{-1mm}
    \captionof{figure}{Compared to existing scanning strategies, our proposed method is capable of producing content-aware paths with flexible curve start ($\bullet$), alleviating the inherent causal modeling limitation of Mamba. [Key: the input is down-scaled to $4\times4$ for curve generation]}
    \label{fig_one}
\end{figure}
\vspace{-2mm}

\input{sec/1_intro}

\input{sec/3_prelim}
\input{sec/4_methods}

\input{sec/5_exp}

\input{sec/7_Conclusion}

\section*{Acknowledgements and Disclosures}

\paragraph{Funding.}
This work was supported by the Natural Sciences and Engineering Research Council of Canada (NSERC) through a Discovery Grant and by RBC Borealis through the RBC Borealis AI Global Fellowship Award. Computational resources were provided by the Digital Research Alliance of Canada.

\paragraph{Competing Interests.}
The authors declare no competing interests.

\newpage
{
\bibliographystyle{plain}
\bibliography{main.bib}
}

\newpage
\appendix

\input{sec/6_supp}

\end{document}

%% file: sec/0_abstract.tex
\begin{abstract}

Selective state-space models (SSMs) have recently shown strong potential for efficient image restoration. A central difficulty, however, is that a 2D feature map typically needs to be linearized into a 1D token sequence for causal state-space updates, making restoration performance highly sensitive to the imposed scan order. Existing deterministic scans follow space-filling principles but remain fixed and content-agnostic, and can yield fragmented or semantically inconsistent contexts under degradations. 
We propose \textbf{CoScan}, a \textbf{co}ntent-adaptive \textbf{scan}-order design for causal vision SSMs. CoScan learns image-dependent space-filling traversals via differentiable proxy supervision driven by \textbf{scan-quality objectives} that \textit{encourage short-range coherence and causal predictability, while preserving spatial adjacency}. To support hierarchical restoration backbones, we introduce an \textbf{efficient multi-scale scan construction by coarsening a fine-level minimum spanning tree}, producing coherent scans across feature resolutions. In addition, a \textbf{frequency-guided start selection strategy} is designed to \textit{place informative structures early in the sequence prefix to strengthen causal context formation}. Plugged into existing Mamba-based backbones, CoScan yields consistent improvements across several restoration benchmarks with minimal overhead and shows promising results on high-level vision tasks. Code will be released at \url{https://github.com/dongw22/CoScan}.
\end{abstract}

%% file: sec/1_intro.tex
\section{Introduction}
\label{sec:intro}

Image restoration targets the recovery of faithful structures from degraded inputs arising in real-world image acquisition, where visibility loss and motion-induced corruption are particularly common.
Such degradations often suppress fine details and distort boundaries, prompting extensive research into deep learning approaches to mitigate their effects in recent years~\citep{zhou2025lita, dong2026zero, zhou2024glare, dong2026evogs}.

Convolutional architectures~\citep{mao2016image, zhou2023breaking} excel at exploiting local coherence but may struggle to capture long-range dependencies,
while generative approaches~\citep{xia2023diffir, gppllie, dong2026pvrf} often improve perceptual realism via iterative refinement with higher inference cost. Meanwhile, transformer-based models~\citep{vaswani2017attention, dong2024shadowrefiner, dong2024dehazedct, dong2025sgllie} strengthen global interactions at the cost of quadratic attention complexity.

Recently, Mamba~\citep{gu2024mamba}, a selective state-space model (SSM), has gained attention for enabling long-range sequence modeling with \emph{linear complexity} in sequence length, which has led to encouraging results in vision and image restoration~\citep{zhu2024vision, liu2024vmamba, zheng2024u, kong2025efficient, gao2024learning, liu2025xyscannet, tang2025ramir, li2025mair, dong2024ecmamba}.

When applying Mamba to images, a key consideration is \textit{scan order design}, which linearizes a 2D feature map into an ordered 1D token sequence. Since visual cues are spatially distributed and lack a canonical 1D ordering, the imposed scan directly governs how spatial context is presented and aggregated in causal state-space updates.

\begin{figure}[t]
    \setlength{\abovecaptionskip}{1mm}
    \setlength{\parskip}{0pt}
    \centering

    \begin{minipage}{0.9\linewidth}
        \centering

        \includegraphics[width=\linewidth]{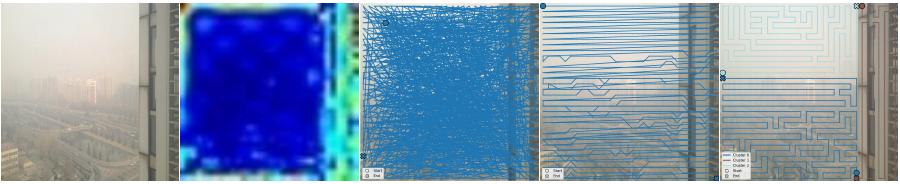}
        \par\nointerlineskip
        \vspace{0.8mm} % Adjust the image-to-label gap here

        \noindent
        \begin{minipage}[t]{0.2\linewidth}
            \centering
            \scriptsize Degraded Input
            \par
        \end{minipage}%
        \begin{minipage}[t]{0.2\linewidth}
            \centering
            \scriptsize High-frequency Map
            \par
        \end{minipage}%
        \begin{minipage}[t]{0.2\linewidth}
            \centering
            \scriptsize MambaIRv2 Scan
            \par
        \end{minipage}%
        \begin{minipage}[t]{0.2\linewidth}
            \centering
            \scriptsize DAMamba Scan
            \par
        \end{minipage}%
        \begin{minipage}[t]{0.2\linewidth}
            \centering
            \scriptsize \textbf{CoScan (Ours)}
            \par
        \end{minipage}
        \par
    \end{minipage}

    \caption{Comparisons on adaptive scans.
    \,{\color{ForestGreen}\small$\bullet$} denotes our final selected
    curve start. Our content-aware scan works well with degraded inputs,
    avoiding discontinuous jumps in MambaIRv2~\cite{guo2025mambairv2}
    and repeated visits in DAMamba~\cite{li2025damamba}. Furthermore,
    our developed frequency-guided start selection strategy
    (Sec.~\ref{section_method_start}) performs well in prefix organization
    by effectively selecting informative regions.}
    \label{fig_frequency_top}
    \vspace{-4mm}
\end{figure}

As shown in Fig.~\ref{fig_one}, most existing Mamba-based vision restoration methods employ a range of scan geometries, including Z-shaped row-wise traversals (Fig.~\ref{fig_one}\textcolor{red}{a}), S-shaped (serpentine) scans (Fig.~\ref{fig_one}\textcolor{red}{b}, Fig.~\ref{fig_one}\textcolor{red}{d}), window-based traversals (Fig.~\ref{fig_one}\textcolor{red}{c}), diagonal sweeps (Fig.~\ref{fig_one}\textcolor{red}{e}), and the Hilbert curve (Fig.~\ref{fig_one}\textcolor{red}{f}).
Despite their different geometries, these scans share a common intent that aligns with the space-filling-curve (SFC) principle: constructing a traversal over the image grid that visits each spatial location exactly once while encouraging spatial locality. However, being fixed and content-agnostic, they may still introduce frequent boundary crossings or long-range jumps, especially under image degradations, leading to fragmented and semantically inconsistent contexts. Recent studies \cite{guo2025mambairv2, li2025damamba} have started to investigate dynamic scanning mechanisms based on deformable convolution or prompt learning. However, these approaches exhibit \textbf{notable limitations}: MambaIRv2~\cite{guo2025mambairv2} contains \textit{\textbf{significant jumps that deteriorate the locality of adjacent tokens}}, whereas DAMamba~\cite{li2025damamba} leads to \textit{\textbf{repeated visits}} to specific tokens and shows only \textit{\textbf{limited deviation}} from standard row-wise scanning~\cite{liu2024vmamba} (see Fig.~\ref{fig_frequency_top}).

This perspective motivates revisiting SFC-based scan design as a principled way to impose a context-aware 2D-to-1D ordering for vision SSMs. NeuralSFC~\citep{wang2022neural} enables content-adaptive traversals by learning dual-graph weights from the input (Fig.~\ref{fig_one}\textcolor{red}{g}). \textbf{However, directly adopting such learned SFCs for vision SSMs remains nontrivial}. 

First, the learning objective of NeuralSFC is \textbf{not explicitly aligned} with the requirements of causal state-space propagation, so the induced ordering may still \textit{allocate early sequence positions to less informative regions} (see Fig.~\ref{fig_one}\textcolor{red}{g} and Fig.~\ref{fig_curve_comapre}\textcolor{red}{b}). Second, existing learned SFCs are \textbf{largely single-scale}, which limits their direct deployment in existing hierarchical networks~\cite{gu2024mamba,mohamhaze,li2025mair,dong2026towards}.

To bridge these gaps, we propose \textbf{CoScan}, a \textbf{co}ntent-adaptive \textbf{scan}-order design for causal state-space models.
CoScan generates space-filling traversals with multi-scale consistency and adaptive start control, and can be plugged into existing Mamba-based backbones for improved performances.

We summarize our contributions as following:
\begin{itemize}[topsep=2pt,itemsep=1.5pt,parsep=0pt,partopsep=0pt,leftmargin=*]
    \item Targeting Mamba's causal state-space modeling, we introduce a differentiable learning framework for content-adaptive space-filling scans, using a proxy evaluator and \textbf{scan-quality objectives} to improve \textit{short-range coherence and causal predictability} under degradations.
\item We propose an \textbf{efficient multi-scale scan construction via graph-based coarsening}, enabling consistent traversals across feature resolutions without regenerating curves at each scale.
    \item We design a \textbf{lightweight frequency-guided start selection strategy} to place informative regions early in the sequence prefix. \textit{Consistent gains} are observed when plugging CoScan into standard Mamba-based backbones \textit{without introducing extra computation overhead}.
\end{itemize}

%% file: sec/3_prelim.tex
\section{Preliminaries}

\textbf{Mamba}~\citep{gu2024mamba} is a selective state space sequence model that updates a latent state recurrently with token-wise, input-adaptive parameters. 
Given an input sequence $\{x_k\}_{k=1}^{L}$, Mamba computes
\begin{equation}
h_k=\bar{\mathbf{A}}_k h_{k-1}+\bar{\mathbf{B}}_k x_k, \: y_k=\bar{\mathbf{C}}_k h_k,
\label{eq_mamba_tokenwise}
\end{equation}
where $h_k$ and $y_k$ denote the latent state and output, and $\bar{\mathbf{A}}_k$, $\bar{\mathbf{B}}_k$, and $\bar{\mathbf{C}}_k$ are generated conditioned on the input token.

\textbf{A space-filling curve} (SFC) corresponds to a Hamiltonian traversal on the image grid, producing a 1D ordering that visits every pixel exactly once \citep{dafner2000context}, as illustrated in Fig.~\ref{fig_teaser}~\textcolor{red}{I(a)}.

The construction starts from a circuit cover of an $H\times W$ grid, denoted as $\mathcal{G}$, formed by disjoint $2\times2$ circuits $\{C_i\}$, whose vertices are pixel locations. A dual graph $\mathcal{G^{\prime}}$ is then constructed, where each node represents circuits and each edge connects adjacent circuit pairs that can be merged. For each neighboring pair $(C_i,C_j)$, a merge cost $W(C_i,C_j)$ is evaluated by exchanging two intra-circuit edges $(e, f)$ with two cross-circuit edges $(u, v)$ as follows:
\begin{equation}
  W(C_i,C_j)=|u|+|v|-|e|-|f|,
  \label{eq_merge_cost_dafner}
\end{equation}
where $|\cdot|$ denotes the absolute difference of pixel intensities at the two vertices of an edge in $\mathcal{G}$. Upon these weights, a minimum spanning tree (MST) $\mathcal{T}$ is computed on $\mathcal{G}'$ (Fig.~\ref{fig_teaser}~\textcolor{red}{I(c)}), and merging circuits along $\mathcal{T}$ yields a single Hamiltonian circuit (Fig.~\ref{fig_teaser}~\textcolor{red}{I(d)}). Finally, cutting the Hamiltonian circuit at an arbitrary breakpoint yields the SFC order.

%% file: sec/4_methods.tex
\section{Method}
\label{sec_method}

\input{sec/4_1_Motivation}

\input{sec/4_2_CASFC}

\begin{figure*}[t]
    \setlength{\abovecaptionskip}{0.05in}
    \centering\setlength{\baselineskip}{0mm}
    \vspace{-2mm}
    \includegraphics[width=\linewidth]{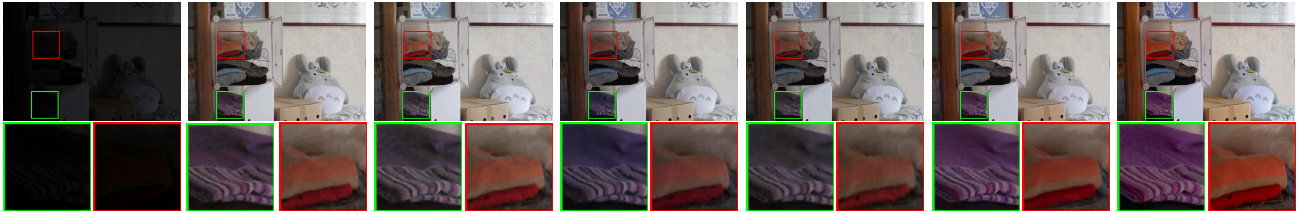}
    \begin{minipage}[b]{0.136\linewidth}
    \centering
    \scriptsize{Input}
    \end{minipage}
    \begin{minipage}[b]{0.136\linewidth}
    \centering
    \scriptsize{UCMNet}
    \end{minipage}
    \begin{minipage}[b]{0.136\linewidth}
    \centering
    \scriptsize{EAMamba}
    \end{minipage}
    \begin{minipage}[b]{0.136\linewidth}
    \centering
    \scriptsize{MambaIR}
    \end{minipage}
    \begin{minipage}[b]{0.136\linewidth}
    \centering
    \scriptsize{MaIR}
    \end{minipage}
    \begin{minipage}[b]{0.136\linewidth}
    \centering
    \scriptsize{\textbf{CoScan + MaIR}}
    \end{minipage}
    \begin{minipage}[b]{0.136\linewidth}
    \centering
    \scriptsize{GT}
    \end{minipage}
    \includegraphics[width=\linewidth]{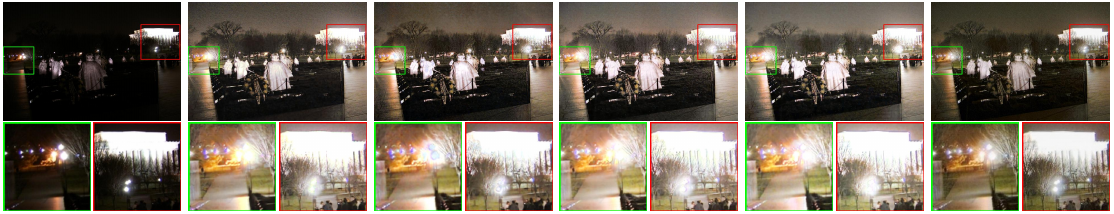}
    \begin{minipage}[b]{0.16\linewidth}
    \centering
    \scriptsize{Input}
    \end{minipage}
    \begin{minipage}[b]{0.16\linewidth}
    \centering
    \scriptsize{Restormer}
    \end{minipage}
    \begin{minipage}[b]{0.16\linewidth}
    \centering
    \scriptsize{EAMamba}
    \end{minipage}
    \begin{minipage}[b]{0.16\linewidth}
    \centering
    \scriptsize{MambaIR}
    \end{minipage}
    \begin{minipage}[b]{0.16\linewidth}
    \centering
    \scriptsize{MaIR}
    \end{minipage}
    \begin{minipage}[b]{0.16\linewidth}
    \centering
    \scriptsize{\textbf{CoScan + MaIR}}
    \end{minipage}
    
    \caption{Visual comparisons on the LLIE task. Compared with prior methods that may yield over-smoothing, color shifts, and residual noise or over-amplify bright areas, our method produces visually pleasing results with clearer structures, finer details and more faithful color and brightness.}
    \label{fig_vis_lolv1}
    \vspace{-6mm}
\end{figure*}

\begin{minipage}[!t]{0.51\linewidth}
\centering
    \footnotesize
	\setlength\tabcolsep{1pt}
    \setlength{\abovecaptionskip}{2mm}
    \renewcommand{\arraystretch}{1.15}
    \captionof{table}{Quantitative comparisons (deblurring task).}  
	\resizebox{0.9\textwidth}{!}{
        \centering
		\begin{tabular}{l|ccc|ccc}
			\toprule[0.15em]
			\multirow{2}{*}{Methods}      & \multicolumn{3}{c|}{GoPro} & \multicolumn{3}{c}{HIDE}  \\  
           & PSNR$\uparrow$ & SSIM$\uparrow$ & LPIPS$\downarrow$ & PSNR$\uparrow$ & SSIM$\uparrow$ & LPIPS$\downarrow$  \\ \midrule[0.15em]
            HINet~\cite{hinet}   &30.11 &0.895 &0.142 &28.17 &0.869 &0.175  \\   %confirmed
            
            NAFNet~\cite{nafnet}   &28.94 &0.876 &0.210 &26.87 &0.842 &0.266  \\   %confirmed
            SwinIR~\cite{liang2021swinir}   &25.79 &0.786 &0.228 &24.02 &0.756 &0.298  \\  %confirmed

			UFormer~\cite{uformer}    &31.95 &0.921 &0.115 &30.08 &0.900 &0.136  \\ %confirmed
			
			Restormer~\cite{restormer}    &32.02 &0.920 &0.114 &30.26 &0.903 &0.134   \\ 
            
            BioIR~\cite{bioir}  & 30.37  &0.902   &0.138   &28.85 &0.877 &0.168   \\%confirmed

            UCMNet~\cite{ucmnet} &32.17 &0.920 & 0.115 & 30.11 & 0.902 & 0.136   \\%confirmed
            [1mm]
            \hline
            \noalign{\vspace{1mm}}
            MambaIR~\cite{guo2024mambair}   &27.70 &0.849 &0.239 &26.10 &0.824 &0.292 \\  %confirmed
            \textbf{CoScan + MambaIR}    &29.36 &0.892 &0.151 &28.22 &0.871 &0.172\\

            EAMamba~\cite{lin2025eamamba}   &30.15 &0.900 &0.137 &28.57 &0.881 &0.166 \\  %confirmed
            \textbf{CoScan + EAMamba}    &31.28 &0.916 &0.120 &30.15 &0.904 &0.133 \\

            MaIR~\cite{li2025mair} &\textcolor{blue}{32.97} &\textcolor{blue}{0.928} &\textcolor{blue}{0.102} &\textcolor{blue}{30.42} &\textcolor{blue}{0.911} &\textcolor{blue}{0.127}   \\
            \textbf{CoScan + MaIR}   &\textcolor{red}{33.55} &\textcolor{red}{0.941} &\textcolor{red}{0.081} &\textcolor{red}{31.76} &\textcolor{red}{0.929} &\textcolor{red}{0.104}   \\
			\bottomrule[0.15em]
	\end{tabular}}
	\label{tab:quantitative_deblurring}
\end{minipage}\hfill
\begin{minipage}[!t]{0.47\linewidth}
\centering
    \footnotesize
	\setlength\tabcolsep{1pt}
    \setlength{\abovecaptionskip}{2mm}
    \renewcommand{\arraystretch}{1.15}
    \captionof{table}{Quantitative results (raindrop removal).} 
	\resizebox{1\textwidth}{!}{
		\begin{tabular}{l|ccccc}
			\toprule[0.15em]
			\multirow{2}{*}{Methods}    & \multicolumn{5}{c}{Raindrop Dataset} \\  
             & PSNR$\uparrow$ & SSIM$\uparrow$ & LPIPS$\downarrow$ & MANIQA$\uparrow$ & MUSIQ$\uparrow$ \\ \midrule[0.15em]
            HINet~\cite{hinet}    &31.20   &\textcolor{red}{0.924}    &0.060  &0.486   &\textcolor{blue}{71.50}    \\
            NAFNet~\cite{nafnet}    &31.13   &0.921    &\textcolor{blue}{0.059}  &0.484  &71.48    \\   
			UFormer~\cite{uformer}    &27.81   &0.878    &0.109  &0.390   &64.77   \\
			Restormer~\cite{restormer}     &30.86   &0.920    &0.060  &\textcolor{blue}{0.487}   &71.45     \\
            BioIR~\cite{bioir}  &30.75 &0.919 &0.064    &0.480  &70.89     \\
            UCMNet~\cite{ucmnet}   &31.09   &0.921  &0.063  &0.486   &71.18    \\
            [1mm]
            \hline
            \noalign{\vspace{1mm}}
            MambaIR~\cite{guo2024mambair}   &30.23   &0.915    &0.065  &0.479   &71.11  \\
            \textbf{CoScan + MambaIR}    &31.18   &0.920    &0.061  &0.486   &71.49    \\
            EAMamba~\cite{lin2025eamamba}    &30.31   &0.913    &0.065  &0.476   &70.83   \\
            \textbf{CoScan + EAMamba}    &\textcolor{blue}{31.25}   &0.919    &0.061  &0.485   &71.42 \\
            MaIR~\cite{li2025mair}  &30.74   &0.918    &0.060  &0.481  &71.08    \\
            \textbf{CoScan + MaIR}      &\textcolor{red}{32.45}   &\textcolor{blue}{0.923}    &\textcolor{red}{0.059}  &\textcolor{red}{0.488}   &\textcolor{red}{71.57}   \\
			\bottomrule[0.15em]
	\end{tabular}}
	\label{tab:quantitative_raindrop}
\end{minipage}

%% file: sec/4_1_Motivation.tex
\subsection{Motivation: Scan-Order Design under Causality}
%Causal Modeling Challenge of Mamba
\label{sec_method_motivation}
Mamba performs causal state-space modeling via the recurrent update in Eq.~\ref{eq_mamba_tokenwise}, where $y_k$ depends only on the current token and its preceding context, \textit{i.e.}, $\{x_t\}_{t\le k}$. Meanwhile, the influence of historical tokens decays exponentially with respect to the sequence lag $(k-t)$, so that earlier tokens contribute progressively less to the current update. While such a locality-biased behavior is well-suited to text sequences, where semantically relevant context is often concentrated in nearby tokens, in vision it becomes highly sensitive to \textit{\textbf{how a 2D grid is linearized into 1D sequence}}, since the scan order determines whether spatially correlated pixels are placed within an effective neighborhood. This motivates considering space-filling curves (SFCs) for scan design, a locality-preserving 2D-to-1D linearization that has received limited attention in Vision Mamba. Nevertheless, using an off-the-shelf SFC as the scan order poses \textit{\textbf{the following challenges}} in image restoration:
\begin{itemize}[topsep=2pt,itemsep=1.5 pt,parsep=0pt,partopsep=0pt,leftmargin=*]
    \item A continuous, locality-preserving traversal on the 2D grid does not necessarily yield \textbf{\emph{short-lag coherence}} in the 1D order: semantically correlated pixels can still be separated by long sequence gaps, especially when degradations corrupt local cues used to construct the traversal.

    \item Hierarchical restoration backbones require scan orders \textit{\textbf{at multiple resolutions}}, yet regenerating a content-aware curve independently per scale is costly and may introduce undesirable cross-scale inconsistency.

    \item  Causality makes early tokens influence more subsequent updates, but off-the-shelf SFCs do not control \textbf{\textit{how early informative regions appear}} in the sequence. As a result, salient structures may be visited too late to be effective.

\end{itemize}

\begin{figure*}[t]
\setlength{\abovecaptionskip}{1mm}
\setlength{\parskip}{0mm} 
\setlength{\baselineskip}{0mm} 
\centering
\begin{minipage}[c]{0.32\linewidth}
    \centering
    \includegraphics[width = 1\textwidth]{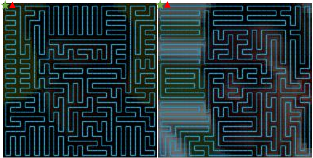}
    \setlength{\parskip}{0mm} 
    \setlength{\baselineskip}{0mm}
    \phantomsubcaption\label{Dafner}
    \scriptsize{(a) Dafner~\citep{dafner2000context}}
%    \caption*{Dafner et.al}
\end{minipage}
\begin{minipage}[c]{0.32\linewidth}
    \centering
    \includegraphics[width = 1\textwidth]{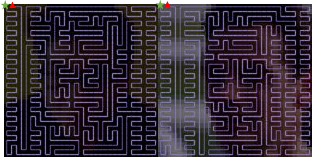}
    \setlength{\parskip}{0mm} 
    \setlength{\baselineskip}{0mm}
    \phantomsubcaption\label{NeuralSFC}
    \scriptsize{(b) NeuralSFC~\citep{wang2022neural}}
\end{minipage}
\begin{minipage}[c]{0.32\linewidth}
    \centering
    \includegraphics[width = 1\textwidth]{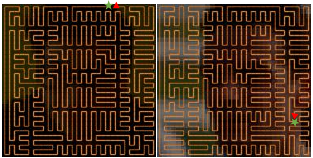}
    \setlength{\parskip}{0mm} 
    \setlength{\baselineskip}{0mm}
    \phantomsubcaption\label{Ours}
    \scriptsize{(c) \textbf{Ours}}
\end{minipage}
\caption{Dafner method and NeuralSFC struggle with degraded inputs. In contrast, our content-aware curves are capable of traversing informative regions before transiting to the background. In addition, the adaptive start selection algorithm allows informative regions to be maximally exploited, thereby alleviating the challenges of causal state-space modeling in Mamba. [Key: {\color{ForestGreen}\ding{72}}: Start; {\color{red}\ding{115}}: End]}
\label{fig_curve_comapre}
\vspace{-6mm}
\end{figure*}

Accordingly, CoScan introduces a tailored content-aware SFC generation pipeline with scan-quality objectives (Sec.~\ref{section_method_CMSFC}), adaptive curve scaling (Sec. \ref{section_method_scaling}), and an informative-region-aware start selection mechanism (Sec.~\ref{section_method_start}), and integrates them into Mamba backbones for multiple restoration tasks.

%% file: sec/4_2_CASFC.tex
\subsection{Content-aware SFC via Differentiable Optimization}
\label{section_method_CMSFC}
As shown in Fig.~\ref{fig_curve_comapre}, under low-light and low-resolution degradations (with noise/blur), the induced SFC may traverse background areas before adequately covering informative regions, weakening the prefix context for causal modeling. Moreover, the MST-based construction is non-differentiable, which prevents direct optimization of scan quality with gradient-based learning. We therefore follow the parameterization of NeuralSFC~\citep{wang2022neural} and represent an SFC by the dual-graph edge weights $W$: 
given an input image $\mathbf{I}$, a weight generator $G$ predicts per-image weights $W_G = G(\mathbf{I})$.
Originally, we introduce a scan-quality objective vector $\boldsymbol{\mathcal{J}}(W,\mathbf{I})$ (Sec.~\ref{sec:OBJ}) and train an evaluator $E$ (Sec.~\ref{sec:train_E}) to provide differentiable supervision by predicting its estimate
$\widehat{\boldsymbol{\mathcal{J}}} = E(W,\mathbf{I})$,
which serves as a proxy objective to optimize $G$ (Sec.~\ref{sec:train_G}).

\subsubsection{Scan-quality Objectives}
\label{sec:OBJ}
For a traversal induced by $W$, we linearize $\mathbf{I}$ into $\{x_i\}_{i=1}^{N}$ ($N=HW$) and compute three objectives:

\textbf{Autocorrelation (AC).}
We measure short-range dependency by the lag-$l$ autocorrelation
\begin{equation}
\mathcal{J}_{\mathrm{AC}}(W,\mathbf{I})
= \rho_l(\mathbf{x})=
\frac{\sum_{i=1}^{HW-l} x_i\,x_{i+l}}{\sum_{i=1}^{HW} x_i^2}.
    \label{eq:obj_ac}
\end{equation}

\textbf{Autoregressive (AR) residual variance ratio.} 
We quantify causal predictability using a linear autoregressive predictor with lag order $p=8$. It predicts $x_i$ from the previous $p$ samples $\{x_{i-k}\}_{k=1}^{p}$ via a linear combination:
%\begin{equation}
$\hat{x}_i=\sum_{k=1}^{p} a_k\,x_{i-k}, i=p+1,\ldots,N,$
%\end{equation}
where $\mathbf{a}=[a_1,\ldots,a_p]^\top$ is estimated from the sample autocorrelations via the Yule--Walker equations (solved using the Durbin--Levinson recursion)~\cite{eshel2003yule, Levinson-Durbin-recursion}. Then, we define the residual sequence $e_i = x_i-\hat{x}_i$ and measure the residual variance ratio as
\begin{equation}
\mathcal{J}_{\mathrm{AR}}(W,\mathbf{I})
=
\frac{\mathrm{Var}\big(\{e_i\}_{i=p+1}^{N}\big)}
{\mathrm{Var}\big(\{x_i\}_{i=1}^{N}\big)+\varepsilon},
\label{eq:obj_ar}
\end{equation}
where $\varepsilon =1e^{-12}$ is a small constant for numerical stability.
Smaller values indicate that the scanned sequence is more predictable from a short causal history.

\textbf{Spatial adjacency preservation (ADJ).}
Let $\mathcal{N}$ be the set of 4-neighborhood pixel pairs in the $H\times W$ grid, $\pi(u)\in\{1,\ldots,N\}$ denotes the position of pixel $u$ along the traversal induced by $W$, and let $w(u,v)$ denote a generator-produced or Dafner-produced appearance-aware weight.
We penalize neighborhood discontinuities by
\begin{equation}
\mathcal{J}_{\mathrm{ADJ}}
= \mathbb{E}_{(u,v)\in\mathcal{N}}
\left[(1-w(u,v)) \left( | \pi(u)-\pi(v) | - 1 \right)\right].
\label{eq_adj_metric}
\end{equation}

\subsubsection{Training the Evaluator}
\label{sec:train_E}
As illustrated in Fig.~\ref{fig_teaser}~\textcolor{red}{II}, for each input image $\mathbf{I}$, we compute two weight configurations: $W_D$ from Dafner~\citep{dafner2000context} and $W_G$ from the generator $G$ (fixed).
Each weight is passed through the non-differentiable MST module to obtain the corresponding traversal and its induced sequence, from which we compute the \emph{ground-truth} objective vector
$\boldsymbol{\mathcal{J}}(W,\mathbf{I})=[-{\mathcal{J}}_{\mathrm{AC}},{\mathcal{J}}_{\mathrm{AR}},{\mathcal{J}}_{\mathrm{ADJ}}]$.
We then train the evaluator to regress these objectives. Using both $W_D$ and $W_G$ exposes $E$ to a diverse range of weight inputs, thereby improving the reliability of the proxy.
\begin{equation}
\min_{E}\ \mathcal{L}_E
=\mathbb{E}_{W\in\{W_D,W_G\}}
\Big[
\big\|E(W,\mathbf{I})-\boldsymbol{\mathcal{J}}(W,\mathbf{I})\big\|_2^2
\Big].
\label{eq:loss_E}
\end{equation}

\begin{figure*}[t]
    \centering
    \setlength{\abovecaptionskip}{0.05in}
    \centering\setlength{\baselineskip}{0mm}
    \includegraphics[width=0.92\textwidth]{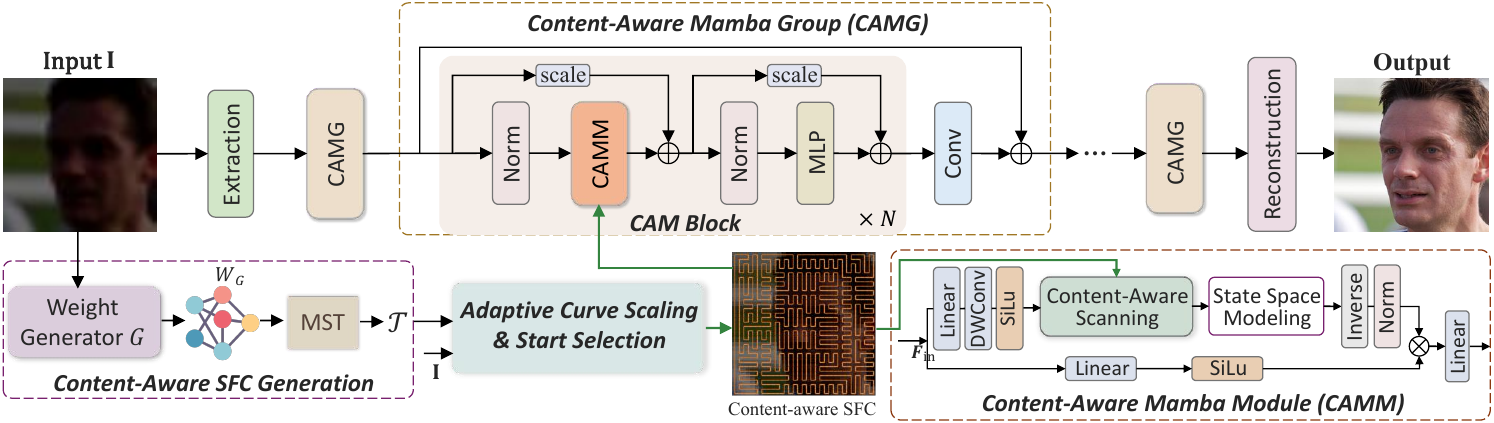}
    \caption{Built upon our CoScan, we introduce a Content-Aware Mamba Module (CAMM) that integrates content-aware serialization into causal state-space modeling in a plug-and-play manner.}
    \label{fig_overall_framework}
    \vspace{-2mm}
\end{figure*}

\subsubsection{Training the Generator}
\label{sec:train_G}
With the evaluator $E$ fixed, we update the generator $G$ by backpropagating through the proxy predictions
$\widehat{\boldsymbol{\mathcal{J}}}(W,\mathbf{I})=E(W,\mathbf{I})$.
We anchor the autocorrelation objective to the Dafner reference by enforcing the predicted AC score of $W_G$ to exceed that of $W_D$ with a margin:
\begin{equation}
\mathcal{L}^{G}_{\mathrm{rank}}
=
\mathrm{softplus}\!\left(
\frac{
\widehat{\mathcal{J}}_{\mathrm{AC}}(W_D,\mathbf{I})
-
\widehat{\mathcal{J}}_{\mathrm{AC}}(W_G,\mathbf{I})
+m}{\tau}
\right),
\label{eq:loss_rank}
\end{equation}
where $m>0$ is the margin and $\tau>0$ is the temperature.
We apply this relative constraint only to AC, since $W_D$ provides a stable reference for autocorrelation, while the AR and ADJ terms are introduced to reflect causal predictability and spatial adjacency preservation and are therefore optimized directly on $W_G$.
This optimization is shown as Eq.~\ref{eq:loss_G}, where $\lambda=0.02,\beta=0.2,\gamma=0.2$ balance the terms.
\begin{multline}
\min_{G}\ \mathcal{L}_G 
=
\mathbb{E}\Big[
\lambda\,\mathcal{L}^{G}_{\mathrm{rank}}
+\big\|
-\widehat{\mathcal{J}}_{\mathrm{AC}}(W_G,\mathbf{I}) - (-1)
\big\|_2 +\beta\,\widehat{\mathcal{J}}_{\mathrm{AR}}(W_G,\mathbf{I})
+\gamma\,\widehat{\mathcal{J}}_{\mathrm{ADJ}}(W_G,\mathbf{I})
\Big].
\label{eq:loss_G}
\end{multline}

\begin{table*}[t]
	\centering
    \footnotesize
	\setlength\tabcolsep{1.5pt}
    \setlength{\abovecaptionskip}{2mm}
    \renewcommand{\arraystretch}{1.05}
    \caption{Quantitative comparisons on the low-light task. Methods are evaluated on LOL~\cite{LOLv1} test set and other four real-world datasets (MEF~\cite{MEF}, DICM~\cite{DICM}, LIME~\cite{LIME}, and NPE~\cite{NPE}).} 
	\resizebox{1\textwidth}{!}{
		\begin{tabular}{l|ccccc|cc|cc|cc|cc|cc}
			\toprule[0.15em]
			\multirow{2}{*}{Methods}    & \multicolumn{5}{c|}{LOL} & \multicolumn{2}{c|}{MEF}  &\multicolumn{2}{c|}{DICM}   &\multicolumn{2}{c|}{LIME} &\multicolumn{2}{c|}{NPE} & \multicolumn{2}{c}{Average (Real-world)}\\  
             & PSNR$\uparrow$ & SSIM$\uparrow$ & LPIPS$\downarrow$ & MANIQA$\uparrow$ & MUSIQ$\uparrow$& CLIP-IQA$\uparrow$ & NIQE$\downarrow$ & CLIP-IQA$\uparrow$ & NIQE$\downarrow$ & CLIP-IQA$\uparrow$ & NIQE$\downarrow$ & CLIP-IQA$\uparrow$ & NIQE$\downarrow$ & CLIP-IQA$\uparrow$ & NIQE$\downarrow$  \\ \midrule[0.15em]
            HINet~\cite{hinet}   & 22.4729   &0.8453   &0.1252 &0.6343  &70.7742   &0.444  &3.85 &0.480 &3.51 &0.431 &\textcolor{blue}{4.14} &0.401 &3.98 &0.463 &3.66   \\
            
            NAFNet~\cite{nafnet}   & 22.7489  &0.8264    &0.1374  &0.4020   &64.8988      &0.449  &3.61 &0.430 &3.98  &0.420 &4.50 &0.373 &4.61  &0.427 &4.01   \\

            SwinIR~\cite{liang2021swinir}  & 17.6107   &0.7377   &0.2591 &0.6072  &59.2487 &0.402  &3.43 &0.436 &3.85 &0.436 &4.63 & 0.329&4.06  &0.422 &3.87  \\

			UFormer~\cite{uformer}   &18.4094   &0.7369   &0.3149 &0.5228  &47.9613   &0.354 &3.89 &0.330 &3.66 &0.336 &4.55 & 0.279&4.02  &0.330 &3.81  \\
			
			Restormer~\cite{restormer}  &23.5068 &0.8379  &0.1225   &0.6255  & 67.9454   &0.484  &3.62 &0.503 &3.58 &0.459 &4.84 &0.367 &3.93 &0.485 &3.73    \\
            
            DGUNet~\cite{dgunet} & 23.1642  & 0.8251   &0.1532 & 0.6020  &66.1795 &0.461  &3.84 &0.493 &3.91 &0.442&4.67 & 0.423 &3.89 &0.477  &3.97 \\
            
            BioIR~\cite{bioir} & 23.2547  & 0.8242   &0.1377 & 0.6130  &66.8548 &0.467  &3.75 &0.497 &3.81 &0.445&4.61 & 0.420 &3.86 &0.481  &3.88 \\
            
            UCMNet~\cite{ucmnet} & 23.7032 & 0.8434  &0.1204  &0.6373  &69.3365 &0.481  &3.65 &0.506 &3.59 &0.456&4.78 & 0.406 &3.83 &0.489  &3.73\\
            [1mm]
            \hline
            \noalign{\vspace{1mm}}
            MambaIR~\cite{guo2024mambair}   &22.7181   & 0.8456   &0.1191  &0.6424 & 70.5206 &0.465  &4.13 &0.482 &3.77 &0.427 &6.60 &0.356 &4.26 &0.464 &4.13 \\
            \textbf{CoScan + MambaIR}  &\textcolor{blue}{23.7910}   &\textcolor{red}{0.8557}  &\textcolor{blue}{0.1070}  &\textcolor{red}{0.6482} &\textcolor{red}{71.9071}  &0.518  &3.49 &0.522 &3.57 &0.463 &4.38 &0.402 &3.98 &0.506 &3.66   \\

            EAMamba~\cite{lin2025eamamba}  &22.8205   & 0.8143   &0.1605  &0.5884  &64.1456  &0.489  &3.53 &0.516 &3.58 &0.451 &4.44 &0.435 &3.75 &0.499 &3.66  \\
            \textbf{CoScan + EAMamba} & 23.7134  & 0.8311   &0.1364  &0.6382   &70.4768  &\textcolor{blue}{0.533}  &\textcolor{blue}{3.40} &\textcolor{red}{0.559} &3.49 &\textcolor{blue}{0.484}  &4.21 &\textcolor{red}{0.459} &\textcolor{red}{3.68}  &\textcolor{blue}{0.539} &\textcolor{blue}{3.55} \\

            MaIR~\cite{li2025mair}  &23.2925   &0.8344    &0.1355  & 0.5884  &65.1757 &0.498  &3.51 &0.496 &\textcolor{blue}{3.45} &0.442 & 4.92& 0.405&3.85 &0.484 &3.63   \\
            \textbf{CoScan + MaIR}  &\textcolor{red}{24.1634} &\textcolor{blue}{0.8537}  &\textcolor{red}{0.1037}  &\textcolor{blue}{0.6436}   &\textcolor{blue}{71.1824} &\textcolor{red}{0.592}  &\textcolor{red}{3.23} &\textcolor{blue}{0.552} &\textcolor{red}{3.43} &\textcolor{red}{0.497}  &\textcolor{red}{3.97} &\textcolor{blue}{0.438} &\textcolor{blue}{3.76} &\textcolor{red}{0.544} &\textcolor{red}{3.47}  \\

			\bottomrule[0.15em]
	\end{tabular}}
	\label{tab:quantitative_lowlight}
	\vspace{-4mm}
\end{table*}

\subsection{Multi-scale Scan Scaling}
\label{section_method_scaling}

To facilitate incorporating our generated curves into existing Mamba networks with the U-Net backbone, we need to downscale or upscale the original curve adaptively to make it aligned with the dimension of the feature map in each scale. We discuss SFC downscaling in this section. Intuitive strategies commonly show limitations: (1) Directly downscaling an SFC leads to discontinuities (dead-ends or abrupt jumps) along the path (see Appendix for details); (2) Generating the SFC at each U-Net scale significantly increases computation.

To enable a feasible incorporation of SFC into Mamba, we propose to downscale the minimum spanning tree (MST) constructed from the original input $\mathbf{I}_{in}$ across a set of target scales $\{\mathcal{S}\}$ in Algorithm \ref{alg:sfc_downscaling}. Rather than re-computing a new SFC for each scale, we manipulate the original fine-grained MST $\mathcal{T}$ directly to preserve global structure from the original curve.

\vspace{-2mm}
\begin{wrapfigure}[21]{r}{0.5\linewidth}
\captionsetup{type=algorithm, labelfont=bf}
\setlength{\intextsep}{0pt}
\hrule height 0.75pt
\vspace{0.5mm}
\caption{Multi-scale SFC Downscaling}
\vspace{-2mm}
\hrule
\label{alg:sfc_downscaling}
\footnotesize
\begin{algorithmic}[1]
\STATE \textbf{Input:} Fine-level MST $\mathcal{T}=(\mathcal{V},\mathcal{E})$ on an $H \times W$ grid, scales $\mathcal{S}=\{s_1,\dots,s_L\}$
\STATE \textbf{Output:} Coarse MSTs $\{\mathcal{T}_s\}_{s\in\mathcal{S}}$, where $\mathcal{T}_s=(\mathcal{V}_s,\mathcal{E}_s)$
\FOR{each $s \in \mathcal{S}$}
    \STATE $H_s \leftarrow \lfloor H/s \rfloor,\;\; W_s \leftarrow \lfloor W/s \rfloor$
    \STATE $\mathcal{V}_s \leftarrow \{(r,c)\mid 0\le r < H_s,\;0\le c < W_s\}$, \quad $\mathcal{B}_s \leftarrow \emptyset$
    \FOR{each edge $(u,v)\in\mathcal{E}$ with weight $w(u,v)$}
        \STATE $a \leftarrow \big(\lfloor u_r/s \rfloor,\lfloor u_c/s \rfloor\big)$, \quad
               $b \leftarrow \big(\lfloor v_r/s \rfloor,\lfloor v_c/s \rfloor\big)$
        \IF{$a=b$}
            \STATE \textbf{continue} \hfill \COMMENT{ignore intra-block edges}
        \ENDIF
        \STATE $e \leftarrow (\min(a,b),\max(a,b))$
        \STATE $\mathcal{B}_s[e] \leftarrow \mathcal{B}_s[e] \cup \{w(u,v)\}$
    \ENDFOR
    \STATE Initialize $G_s=(\mathcal{V}_s,\emptyset)$
    \FOR{each coarse edge $e=(p,q)$ in $\mathcal{B}_s$}
        \STATE $\bar{w}_s(p,q) \leftarrow \mathrm{MedianPool}\big(\mathcal{B}_s[e]\big)$
        \STATE $\mathcal{E}_s \leftarrow \mathcal{E}_s \cup \{(p,q,\bar{w}_s(p,q))\}$
    \ENDFOR
    \STATE $G_s \leftarrow (\mathcal{V}_s,\mathcal{E}_s)$
    \STATE $\mathcal{T}_s \leftarrow \mathrm{MST}(G_s)$
\ENDFOR
\STATE \textbf{return} $\{\mathcal{T}_s\}_{s\in\mathcal{S}}$
\end{algorithmic}
\hrule
\end{wrapfigure}

 For each scale, we partition $\mathcal{T}$ into coarse-blocks corresponding to each downscale factor $s$. All intra-block nodes and edges are collapsed into a single representative node, while inter-block edges are merged through median pooling. This pooled representation is then mapped onto a coarse graph. Finally, we apply an MST algorithm to yield a coarsened MST corresponding to the target scale. The generated coarse MSTs and SFCs are presented in Fig.\ref{fig_teaser}~\textcolor{red}{III}, demonstrating satisfactory curve structure preservation.

\subsection{Frequency-guided Start Selection}
\label{section_method_start}

To better address the open question of how to prioritize important image regions in a causal sequence model, we propose a frequency-domain start selection scheme to \textit{\textbf{determine where the scan should begin}}. Specifically, we first derive an input-dependent frequency saliency map by extracting the high-frequency component in the Fourier domain and normalizing it to $[0,1]$ (see Fig.~\ref{fig_frequency_top} column 2). We then unfold this 2D map into a 1D feature sequence $\mathbf{f}=\{f_t\}_{t=1}^{N}$ by indexing pixels along the SFC, and we partition $\mathbf{f}$ into $K$ contiguous segments using an optimal dynamic-programming formulation that minimizes the within-segment squared error. Each segment is further scored by its average frequency weight $\mu_k$, weighted by its length $n_k$:
\begin{equation}
\label{eq:segment_score}
    s_k=\mu_k\cdot\left(n_k\right)^{\alpha},
\end{equation}
where $\alpha \ge 0$ controls the preference for longer segments. The highest-scoring segment is chosen as the structure-rich interval, and the SFC order is rotated to start from its first occurrence. This interval-level selection reduces sensitivity to degradation-induced frequency distortions, particularly isolated noisy peaks. Fig.~\ref{fig_teaser}~\textcolor{red}{IV} and Fig.~\ref{fig_frequency_top} show that this scheme effectively places an informative interval at the sequence prefix, allowing Mamba to better exploit salient regions under causal modeling.

\subsection{Enhancing Mamba with SFCs}
\label{section_method_enhance}

As show in Fig.~\ref{fig_overall_framework}, we replace the default scans of existing Mamba-based IR networks with our content-aware SFC scan (single scan). 
In each CAMM, $\mathbf{F}_{in}$ is unfolded into a 1D sequence $\mathbf{L}_{in}$ according to the SFC permutation. 
We further prepend 8 global tokens to the sequence prefix, where the tokens $\mathbf{L}_g\in\mathbb{R}^{N_g\times C}$ are predicted by a lightweight MLP from the input feature map: %\in\mathbb{R}^{T_g\times C}, where $T_g$ denotes the number of global tokens and $C$ is the feature dimension. 
\begin{equation}
    \mathbf{L}_g=\mathrm{MLP}\!\left(\mathrm{Pool}(\mathbf{F}_{in})\right),\quad
    \hat{\mathbf{L}}_{in}=[\mathbf{L}_g;\,\mathbf{L}_{in}],
\end{equation}
where $N_g$ denotes the number of global tokens and $C$ is the feature dimension.
The augmented sequence $\hat{\mathbf{L}}_{in}$ is then processed by the state space modeling layers to produce $\hat{\mathbf{L}}_{out}$. 
Before folding back to the 2D domain, we discard the output global tokens and recover the sequence:
\begin{equation}
    \hat{\mathbf{L}}_{out}=\mathrm{SSM}(\hat{\mathbf{L}}_{in}),\quad
    \mathbf{L}_{out}=\hat{\mathbf{L}}_{out}[N_g:],
\end{equation}
which is finally folded back to $\mathbf{F}_{out}$ via the inverse permutation. Notably, CoScan is trained using the same optimization objective as existing Mamba-based IR models.

%% file: sec/5_exp.tex
\section{Experiments}

\subsection{Experimental Setups}
\label{sec_exp_setup}

\textbf{Datasets and Metrics.} We evaluate our CoScan on four image restoration tasks, \textit{i.e.}, low-light image enhancement (LLIE), image deblurring, raindrop removal, image dehazing and super-resolution. For LLIE, all methods are re-trained on LOL \citep{LOLv1} training set and are evaluated on LOL testing samples, along with four additional real-world low-light datasets (MEF \citep{MEF}, DICM \citep{DICM}, LIME \citep{LIME}, and NPE \citep{NPE}). For image deblurring, we train on the GoPro \cite{nah2017deep} training set (2103  blurry-clean image pairs), and evaluate deblurring performance on 1111 and 2045 samples from the GoPro and HIDE \citep{shen2019human} dataset, respectively. In terms of raindrop removal and image dehazing, all methods are trained from scratch and evaluated on Raindrop \citep{qian2018attentive} and RESIDE-6K \citep{qin2020ffa} datasets using their official train-test splits. Following \cite{guo2024mambair, liang2021swinir, li2025mair}, DF2K and DIV2K are employed as the training data for super-resolution and Set5, Set14, B100, Urban100, and Manga 109 are used for testing. We follow \cite{zhang2017learning, zhang2018image} to generate low-resolution images using bicubic interpolation. We utilize Peak Signal-to-Noise Ratio (PSNR) and Structural Similarity Index Measure (SSIM) \citep{SSIM} to assess pixel-level accuracy, and use Learned Perceptual Image Patch Similarity (LPIPS) \citep{LPIPS} for perceptual quality evaluation. In addition, we adopt Natural Image Quality Evaluator (NIQE \citep{NIQE}), CLIP-IQA (\citep{clipiqa}), MANIQA (\citep{maniqa}), and MUSIQ (\citep{musiq}) as non-reference IQA metrics. 

\textbf{Baselines.} We implement CoScan upon three Mamba baselines: MambaIR \citep{guo2024mambair}, EAMamba \citep{lin2025eamamba}, and MaIR \citep{li2025mair}. We compare our CoScan with these three the state-of-the-art (SOTA) Mamba baselines and several representative general image restoration backbones: HINet \citep{hinet}, NAFNet \citep{nafnet}, DGUNet \citep{dgunet}), SwinIR \citep{liang2021swinir}, UFormer \citep{uformer}, and Restormer \citep{restormer}. We also compare recent SOTA task-specific approaches in the Appendix. All baselines are re-trained using their official code. Due to page limit, more evidences are provided in the Appendix:  Implementation details; Related works; More image restoration visualizations, etc.

\begin{figure*}[t]
    \setlength{\abovecaptionskip}{0.05in}
    \centering\setlength{\baselineskip}{0mm}
    \vspace{0mm}
    \includegraphics[width=\linewidth]{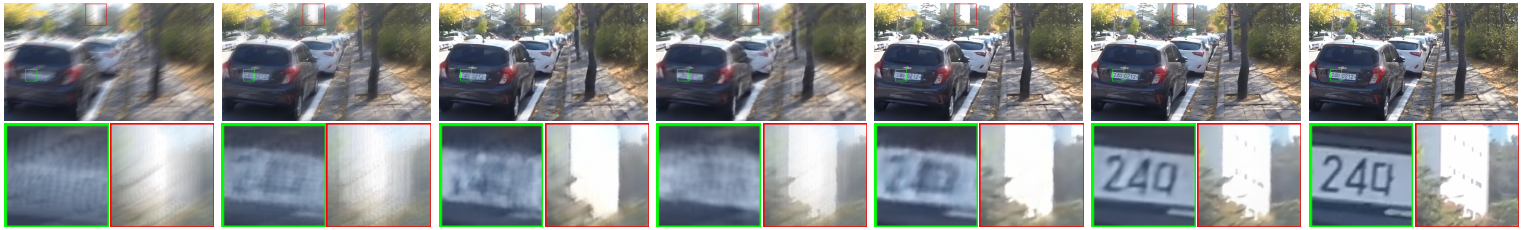}
    \includegraphics[width=\linewidth]{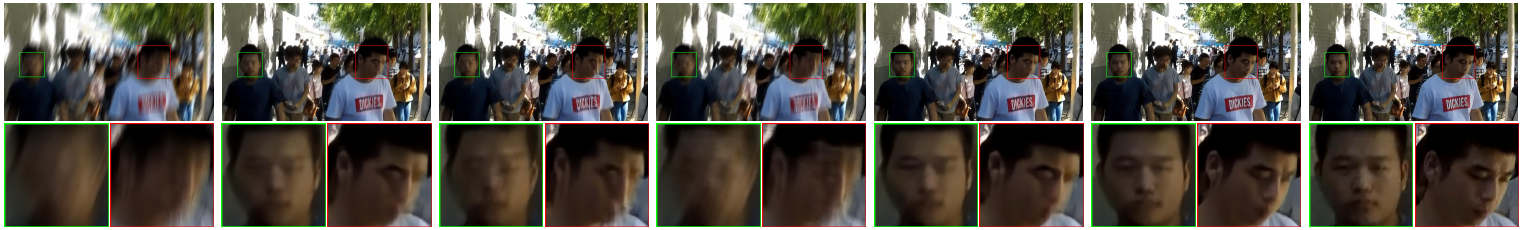}
    \begin{minipage}[b]{0.136\linewidth}
    \centering
    \scriptsize{Input}
    \end{minipage}
    \begin{minipage}[b]{0.136\linewidth}
    \centering
    \scriptsize{UCMNet}
    \end{minipage}
    \begin{minipage}[b]{0.136\linewidth}
    \centering
    \scriptsize{EAMamba}
    \end{minipage}
    \begin{minipage}[b]{0.136\linewidth}
    \centering
    \scriptsize{MambaIR}
    \end{minipage}
    \begin{minipage}[b]{0.136\linewidth}
    \centering
    \scriptsize{MaIR}
    \end{minipage}
    \begin{minipage}[b]{0.136\linewidth}
    \centering
    \scriptsize{\textbf{CoScan + MaIR}}
    \end{minipage}
    \begin{minipage}[b]{0.136\linewidth}
    \centering
    \scriptsize{GT}
    \end{minipage}
    \caption{Visual comparisons (deblurring task). The first row is from GoPro, and the second row from HIDE. Our method consistently recovers sharper structures with fewer artifacts.}
    \label{fig_vis_deblur}
    \vspace{-2mm}
\end{figure*}

\subsection{Quantitative Comparisons with Baselines}

We report quantitative results in Tabs. \ref{tab:quantitative_lowlight}, \ref{tab:quantitative_deblurring}, \ref{tab:quantitative_raindrop},  \ref{tab:quantitative_dehazing}, and \ref{tab:quantitative_super} each corresponding to one image restoration task. Overall best results are highlighted in \textcolor{red}{red} and second-best in \textcolor{blue}{blue}.

\begin{minipage}[t]{0.46\linewidth}
\centering
 %  \footnotesize
       \small   
% \normalsize
	\setlength\tabcolsep{0.8pt}
    \setlength{\abovecaptionskip}{2mm}
    \renewcommand{\arraystretch}{1.2}
    	\captionof{table}{Quantitative comparisons on the dehazing task (RESIDE-6K~\cite{qin2020ffa}). MACs are measured on $128\times128$ patches.} \label{tab:quantitative_dehazing}
        \resizebox{\linewidth}{!}{
%	\resizebox{1\textwidth}{!}{
		\begin{tabular}{l|cc|ccc|ccc}
			\toprule[0.15em]
			\multirow{2}{*}{Methods} & \multicolumn{2}{c|}{Complexity} & \multicolumn{3}{c|}{Indoor} & \multicolumn{3}{c}{Outdoor} \\  
            &Params& MACs & PSNR& SSIM& LPIPS& PSNR& SSIM& LPIPS\\ \midrule[0.15em]
            HINet~\cite{hinet} &88.7 M&42.6 G  &29.86 &\textcolor{red}{0.964}  &0.024 &29.02 &0.962 &0.028  \\
            
            NAFNet~\cite{nafnet} &115.9 M&15.8 G &\textcolor{blue}{30.07}   &0.947 &0.032 &28.77 &0.955 &\textcolor{blue}{0.025} \\

            SwinIR~\cite{liang2021swinir} &11.5 M &188 G  &27.65 &0.939  &0.040 &27.35 &0.951 &0.032   \\

			UFormer~\cite{uformer} &50.4 M&21.4 G  &23.33 &0.926  &0.048 &23.14 &0.919 &0.048  \\
            [1mm]
            \hline
            \noalign{\vspace{1mm}}
            MambaIR~\cite{guo2024mambair} &26.8 M&57.5 G &28.89 &0.958  &0.025 &28.01 &0.954 &0.029 \\
            \textbf{CoScan+MambaIR} &24.9 M& 39.3 G &29.71 &0.962  &\textcolor{red}{0.021} &\textcolor{blue}{29.58} &\textcolor{blue}{0.963} &\textcolor{red}{0.022} \\
            MaIR~\cite{li2025mair} &26.3 M &49.3 G &29.95 &0.960  &0.024 &28.49 &0.956 &0.028    \\
            \textbf{CoScan+MaIR} &23.5 M&35.2 G &\textcolor{red}{30.67} &\textcolor{blue}{0.963}  &\textcolor{blue}{0.022} &\textcolor{red}{31.24} &\textcolor{red}{0.966} &\textcolor{red}{0.022}    \\

			\bottomrule[0.15em]
	\end{tabular}}
\end{minipage}\hspace{0.003\linewidth}    %\hfill
\begin{minipage}[t]{0.536\linewidth}
\centering
   % \footnotesize
%\normalsize
    \small
	\setlength\tabcolsep{0.8pt}
    \setlength{\abovecaptionskip}{2mm}
    \renewcommand{\arraystretch}{1.25}
    \captionof{table}{Quantitative comparisons on classic image super-resolution task ($\times4$). CoScan consistently improves PSNR and SSIM across datasets and baselines, demonstrating the effectiveness of our plug-and-play scanning module for super-resolution.} 
     \resizebox{\linewidth}{!}{
	%\resizebox{1\textwidth}{!}{
		\begin{tabular}{l|cc|cc|cc|cc|cc}
			\toprule[0.15em]
			\multirow{2}{*}{Methods}    & \multicolumn{2}{c|}{Set5} & \multicolumn{2}{c|}{Set14} & \multicolumn{2}{c|}{B100} & \multicolumn{2}{c|}{Urban100} & \multicolumn{2}{c}{Manga109}\\  
             & PSNR& SSIM&  PSNR& SSIM& PSNR& SSIM& PSNR& SSIM& PSNR& SSIM   \\ \midrule[0.15em]
            SwinIR~\cite{liang2021swinir}&32.94 &0.904 &29.12 &0.795 & 27.93 &0.749 &27.43 & 0.825 &32.06 &0.926 \\
            MambaIR~\cite{guo2024mambair}   &32.98  &\textcolor{blue}{0.905} &29.16 &0.796 &28.04 &0.751  &27.73 &0.829 &32.27 &0.927   \\
            \textbf{CoScan+MambaIR}    &33.23  &\textcolor{red}{0.907} &\textcolor{blue}{29.48} &\textcolor{blue}{0.798} &\textcolor{blue}{28.37} &\textcolor{blue}{0.757}  &\textcolor{blue}{27.98} &\textcolor{blue}{0.837} &\textcolor{blue}{32.85} &\textcolor{blue}{0.931}    \\
            MaIR~\cite{li2025mair}  &33.05  &\textcolor{blue}{0.905} &29.24 &0.796 &28.15 &0.753  &27.85 &0.833 &32.59 &0.929    \\
            \textbf{CoScan+MaIR}      &\textcolor{red}{33.34}  &\textcolor{red}{0.907} &\textcolor{red}{29.61} &\textcolor{red}{0.799} &\textcolor{red}{28.51} &\textcolor{red}{0.760}  &\textcolor{red}{28.07} &\textcolor{red}{0.841}  &\textcolor{red}{33.05} &\textcolor{red}{0.934}    \\
			\bottomrule[0.15em]
	\end{tabular}}
	\label{tab:quantitative_super}
\end{minipage}

\begin{figure*}[t]
    \setlength{\abovecaptionskip}{0.05in}
    \centering

    \begin{minipage}{\linewidth}
        \centering
        \includegraphics[width=\linewidth]{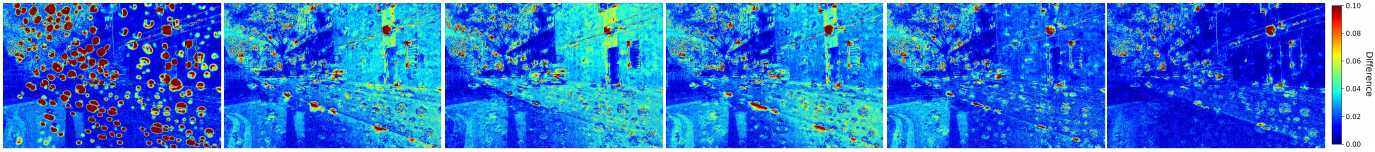}
        \par\nointerlineskip

        % Label positions match the actual panel centers in the PDF.
        \setlength{\unitlength}{\linewidth}
        \begin{picture}(1,0.025)
            \put(0.081595,0.0125){%
                \makebox(0,0){\scriptsize Input}}
            \put(0.241874,0.0125){%
                \makebox(0,0){\scriptsize NAFNet}}
            \put(0.403062,0.0125){%
                \makebox(0,0){\scriptsize EAMamba}}
            \put(0.563523,0.0125){%
                \makebox(0,0){\scriptsize MambaIR}}
            \put(0.724712,0.0125){%
                \makebox(0,0){\scriptsize MaIR}}
            \put(0.884991,0.0125){%
                \makebox(0,0){\scriptsize\textbf{CoScan + MaIR}}}
        \end{picture}
        \par
    \end{minipage}

    \caption{Residual heatmaps between restored results and ground truth
    (raindrop removal). Baselines struggle to reconstruct occluded regions
    behind raindrops, leading to large deviations, whereas our CoScan + MaIR
    shows much less difference.}
    \label{fig_vis_raindrop}
    \vspace{-5mm}
\end{figure*}

\begin{figure*}[t]
    \setlength{\abovecaptionskip}{0.05in}
    \centering

    \resizebox{0.88\textwidth}{!}{%
        \begin{minipage}{\textwidth}
            \centering
            \setlength{\baselineskip}{0mm}

            \includegraphics[width=\linewidth]{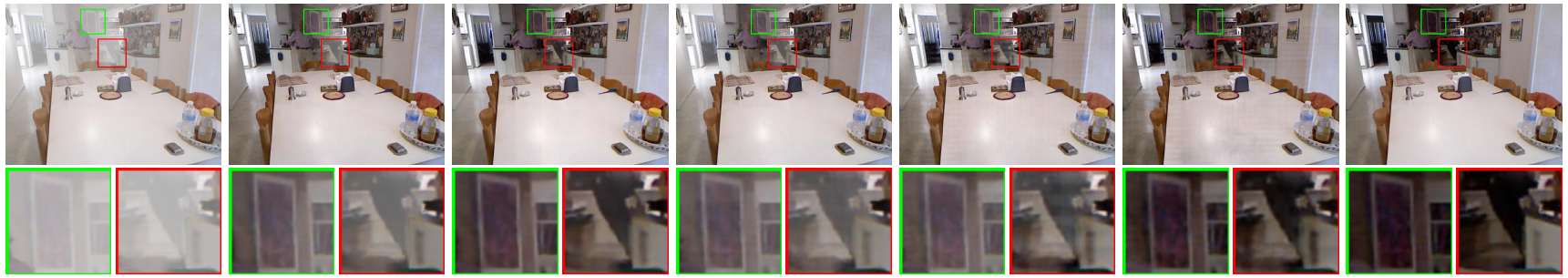}
            \par
            \begin{minipage}[b]{0.137\linewidth}
                \centering
                \scriptsize Input
            \end{minipage}\hfill%
            \begin{minipage}[b]{0.137\linewidth}
                \centering
                \scriptsize NAFNet
            \end{minipage}\hfill%
            \begin{minipage}[b]{0.138\linewidth}
                \centering
                \scriptsize HINet
            \end{minipage}\hfill%
            \begin{minipage}[b]{0.137\linewidth}
                \centering
                \scriptsize MambaIR
            \end{minipage}\hfill%
            \begin{minipage}[b]{0.137\linewidth}
                \centering
                \scriptsize MaIR
            \end{minipage}\hfill%
            \begin{minipage}[b]{0.137\linewidth}
                \centering
                \scriptsize \textbf{CoScan + MaIR}
            \end{minipage}\hfill%
            \begin{minipage}[b]{0.137\linewidth}
                \centering
                \scriptsize GT
            \end{minipage}
            \par

            \includegraphics[width=\linewidth]{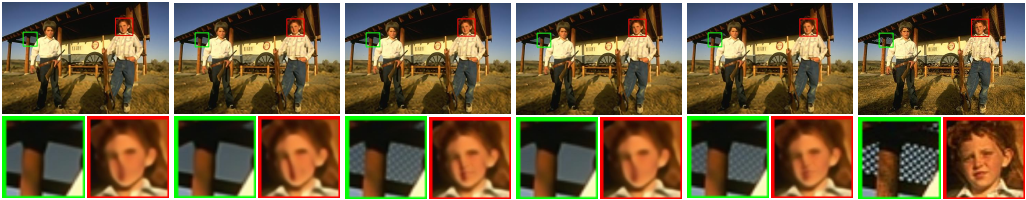}
            \par
            \begin{minipage}[b]{0.16\linewidth}
                \centering
                \scriptsize Input
            \end{minipage}\hfill%
            \begin{minipage}[b]{0.16\linewidth}
                \centering
                \scriptsize MambaIR
            \end{minipage}\hfill%
            \begin{minipage}[b]{0.16\linewidth}
                \centering
                \scriptsize \textbf{CoScan + MambaIR}
            \end{minipage}\hfill%
            \begin{minipage}[b]{0.16\linewidth}
                \centering
                \scriptsize MaIR
            \end{minipage}\hfill%
            \begin{minipage}[b]{0.16\linewidth}
                \centering
                \scriptsize \textbf{CoScan + MaIR}
            \end{minipage}\hfill%
            \begin{minipage}[b]{0.16\linewidth}
                \centering
                \scriptsize GT
            \end{minipage}
            \par
        \end{minipage}%
    }

    \caption{Visual results (dehazing and super-resolution).
    Our method yields visually pleasing outputs.}
    \label{fig_vis_dehazing_super}
    \vspace{-5mm}
\end{figure*}

\textbf{LLIE:} As reported in Tab.~\ref{tab:quantitative_lowlight}, CoScan consistently outperforms across all evaluated datasets for LLIE task. On the LOL dataset, our CoScan variant (built upon MaIR) achieves best overall performance, achieving a PSNR gain of +0.66 dB over the best performing baseline (Restormer). When compared to Mamba baselines, our CoScan yield consistent improvements with a PSNR gain of +1.07 dB (MambaIR), +0.89 dB (EAMamba), and +0.87 dB (MaIR), respectively. Furthermore, our CoScan excel in non-reference metrics. Specifically, CoScan + MaIR achieves the highest average CLIP-IQA and lowest average NIQE scores, surpassing the second-best method by 0.045 and 0.16, respectively.

\textbf{Image Deblurring:} Referring to Tab. \ref{tab:quantitative_deblurring}, our CoScan surpasses all other baselines in both GoPro and HIDE datasets across all fidelity metrics. Notably, CoScan + MaIR outperforms Restormer, the top non-Mamba method, by a gain of +1.53 dB and +1.50 dB in GoPro and HIDE datasets. With our generated content-aware SFC, CoScan + MaIR yields +0.58 dB and +1.34 dB improvements when compared to the MaIR backbone. Such improvements are consistent and stable across Mamba backbone, highlighting the effectiveness and improved generalization from our method.

\textbf{Raindrop Removal, Dehazing and Super-Resolution:} As reported in Tab. \ref{tab:quantitative_raindrop} and Tab. \ref{tab:quantitative_dehazing}, Mamba backbones are inferior to several strong methods (\textit{i.e.}, HINet and NAFNet) on raindrop removal and dehazing tasks. However, CoScan outperforms these non-Mamba methods by enhancing Mamba baselines with our content-aware scans, which effectively alleviates the causal modeling challenge of Mamba and thereby achieving superior performance. Furthermore, Tab. \ref{tab:quantitative_super} indicates that CoScan achieves an average PSNR gain of +0.34 compared to MaIR.

\subsection{Qualitative Results} 
On LOL dataset and real-world low-light scenarios (Fig. \ref{fig_vis_lolv1}), CoScan produces crisper edges while recovering natural color and contrast, yielding more vibrant and detailed scenes than other baselines. For deblurring tasks (Fig. \ref{fig_vis_deblur}), our method performs well in recovering fine details, most notably degraded text and facial features while other baselines struggle to restore legible text and produce smear. For raindrop removal (Fig. \ref{fig_vis_raindrop}), other baselines present significant deviations with GT, whereas CoScan produces more faithful results with low differences. Finally, our methods achieves more appealing results compared to baselines, as visualized in Fig.~\ref{fig_vis_dehazing_super}.

\input{sec/5_exp_abla}

\subsection{Advantages of CoScan over Existing Adaptive Scans}
Existing adaptive scanning approaches (MambaIRv2 and DAMamba) suffer from inherent limitations. As shown in Fig.~\ref{fig_frequency_top}, MambaIRv2 produces significant jumps that break structural continuity, whereas DAMamba incurs repeated visits and exhibits limited deviation from row-wise scans. Our CoScan, however, not only adapts to image content but also introduces a structure-consistent and causality-aligned traversal strategy, leading to improved compatibility with Mamba’s causal modeling process.

%% file: sec/5_exp_abla.tex
\subsection{Ablation Study}

We conduct ablation studies on LOL with the MaIR backbone. The results are reported in Tab.~\ref{tab:abla_1}. 

\noindent
\begin{minipage}[h]{0.54\linewidth}
\textbf{SFC Downscaling, Adaptive Start Selection and Prefix Global Tokens.} We conduct ablations to verify the importance of SFC scaling, start selection and the incorporation of prefix global tokens. The results are listed in Tab.~\ref{tab_abla_first}. The baseline is derived by replacing MaIR scans using our generated SFC. However, the multi-scale scan scaling, adaptive start selection, and prefix global tokens are not deployed in this configuration, and we use an Algorithm~\ref{alg:sfc_downscaling_traditional} for curve downscaling. When progressively adding SFC downscaling (Sec.~\ref{section_method_scaling}), Frequency-guided Start Selection (Sec.~\ref{section_method_start}) and Prefix Global Tokens (Sec.~\ref{section_method_enhance}) to the baseline, we observe +0.45, +0.54, and +0.19 dB improvements. This evidence highlights the effectiveness of our design of multi-scale scan scaling, adaptive start selection, and the introduction of global tokens.
\end{minipage}\hfill
\begin{minipage}[h]{0.44\linewidth}
\vspace{3mm}
\centering
    \footnotesize
    \captionof{table}{Ablation study on LOL dataset. Our final result is highlighted in \textcolor{color2}{purple}.} 
    \label{tab:abla_1}
	\setlength\tabcolsep{3pt}
    \setlength{\abovecaptionskip}{2mm}
    \renewcommand{\arraystretch}{1.0}
    \begin{minipage}[b]{1\linewidth}
    \centering
    %\vspace{-2mm}
    \scriptsize{(a) Ablation on SFC Scaling (SFC-S), Start Selection, and Global Tokens}
    \vspace{4mm}
    \resizebox{1\textwidth}{!}{
		\begin{tabular}{ccccccc}
			\toprule[0.1em]
            \rowcolor{color3} Baseline & SFC-S & Start & Global & PSNR$\uparrow$ & SSIM$\uparrow$ & LPIPS$\downarrow$ \\ \midrule[0.1em]
            \checkmark & & & &22.98 &0.830 &0.135 \\
            \checkmark & \checkmark & & &23.43 &0.836 &0.133 \\
            \checkmark & \checkmark& \checkmark & &23.97 &0.844 &0.114 \\
            \rowcolor{color2} \checkmark  & \checkmark & \checkmark &\checkmark &24.16 &0.854 &0.104 \\
			\bottomrule[0.1em]
	\end{tabular}}
    \label{tab_abla_first}
    \end{minipage}
    \setlength\tabcolsep{0.5pt}
    \begin{minipage}[b]{1\linewidth}
    \centering
    \vspace{-2mm}
    \scriptsize{(b) Ablation on the optimization of CoScan.}
    \vspace{4mm}
    \resizebox{1\textwidth}{!}{
		\begin{tabular}{ccccccccc}
			\toprule[0.1em]
            \rowcolor{color3} Dafner & NeuralSFC &w/o Averaging & Eq.\ref{eq:loss_rank} & Eq.\ref{eq:obj_ar}  &Eq. \ref{eq_adj_metric}  & PSNR$\uparrow$ & SSIM$\uparrow$ & LPIPS$\downarrow$ \\ \midrule[0.1em]
             \checkmark & & & & & &23.44 &0.838 &0.130 \\
              & \checkmark& & & & &23.09 &0.831 &0.133 \\
              & \checkmark&\checkmark & & & &23.43 &0.838 &0.129\\
              &\checkmark &\checkmark &\checkmark & & &23.73 &0.841 &0.118\\
              & \checkmark&\checkmark & \checkmark &\checkmark & &23.88 &0.844 &0.109 \\
              & \checkmark&\checkmark & \checkmark & &\checkmark &23.95 &0.847 &0.107 \\
              \rowcolor{color2}& \checkmark&\checkmark & \checkmark &\checkmark &\checkmark &24.16 &0.854 &0.104 \\
			\bottomrule[0.1em]
	\end{tabular}}
    \label{tab_abla_second}
    \end{minipage}
\end{minipage}

\begin{figure*}[!h]
    \setlength{\abovecaptionskip}{0.05in}
    \centering\setlength{\baselineskip}{0mm}
    \vspace{0mm}
    \includegraphics[width=0.85\linewidth]{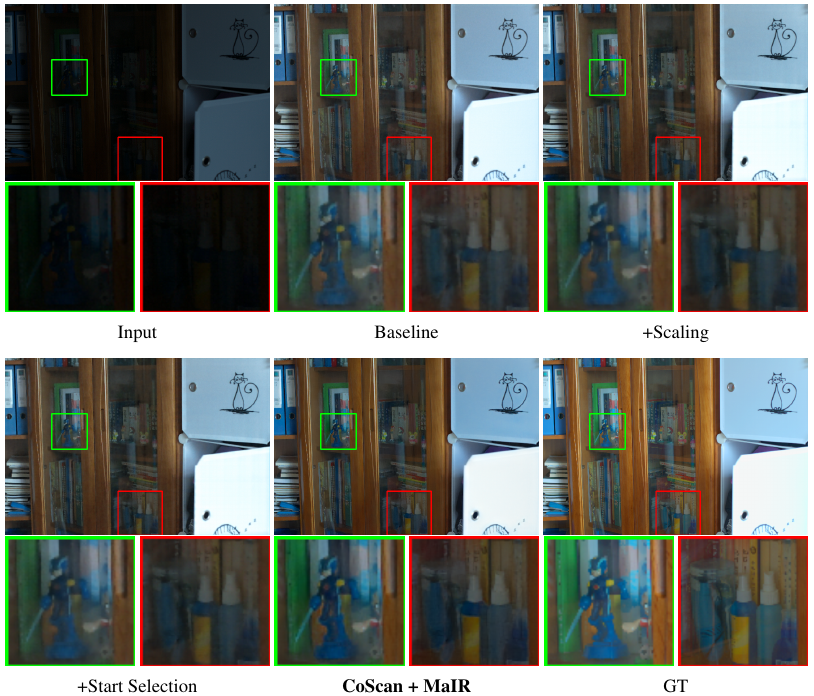}
    \caption{Visualizations for ablation corresponding to Tab.~\ref{tab_abla_first}\textcolor{red}{a} on LOL dataset. Each proposed implementation progressively reduces artifacts and improves color fidelity. More specifically, objects seen through the glass are noticeably brighter, while text appears clearer and more legible.}
    \label{fig_vis_abla_first}
\end{figure*}

\textbf{Content-aware SFC Optimization Process.} We conduct an ablation study to investigate the importance of our generated SFCs, with the deployment of multi-scale scan scaling, adaptive start selection, and global tokens. The results are reported in Tab.~\ref{tab_abla_second}. ``w/o Averaging'' denotes removing the mini-batch weight averaging step, which is unfavorable to generate content-aware curve across images, and this leads to a 0.34 dB improvement in PSNR. With introducing Dafner weight for anchoring autocorrelation optimization, we observe a 0.30 dB PSNR increase and a 0.011 improvement in LPIPS. More importantly, the two objectives introduced in Eqs.~\ref{eq:obj_ar} and \ref{eq_adj_metric} significantly enhance the quality of the SFC, evidenced by the noticeable SSIM improvement (+0.013) on the LOL dataset. Visual comparisons for ablation study are provided in Fig.~\ref{fig_vis_abla_first}.

%% file: sec/7_Conclusion.tex
\section{Conclusion}

We present CoScan, a content-aware scanning framework for Mamba-based image restoration. Our key contributions include (i) a differentiable optimization pipeline for learning high-quality, input-adaptive SFC orders via a proxy evaluator and scan-quality objectives, (ii) a frequency-guided adaptive start selection strategy to better exploit informative regions under causal state-space modeling, and (iii) an efficient multi-scale SFC scaling scheme derived from a fine-grained MST for seamless integration into U-Net backbones. Experiments on multiple restoration tasks demonstrate that integrating our content-aware scans into different Mamba backbones yields consistent and notable improvements.

%% file: sec/6_supp.tex
\section{Implementation Details} 
\label{sec_imple}

We use the Adam optimizer ($\beta_{1}=0.9$, $\beta_{2}=0.99$) and two RTX 2080Ti for all training stages. In the first stage, each training image is cropped offline into multiple $128\times128$ patches without overlap. Following \cite{wang2022neural}, $l$ in Eq.~\ref{eq:obj_ac} is set to 6, and we employ the same initial learning rate and training schedule as in \cite{wang2022neural}. The number of training epochs is set to 100 on LOL and 20 on the other datasets. Based on training stability and empirical results, $m$ and $\tau$ in Eq.~\ref{eq:loss_rank}, as well as $K$ and $\alpha$ in Sec.~\ref{section_method_start}, are set to 0.03, 0.02, 3, and 2.0, respectively.
In the second stage, the training iteration is set to 700K with a batch size of $4$ and image size of $128\times128$. The initial learning rate is set to $10^{-4}$, and is halved at 350K, 500K, 600K, and 650K iterations. Notably, baselines may adopt different training configurations in their official repository, we employ a uniform training pipeline in this work across baselines for fair comparisons. 

\section{Traditional SFC Downscaling}

\begin{figure}[!h]
    \centering
    \setlength{\abovecaptionskip}{2mm}
    \includegraphics[width=0.7\linewidth]{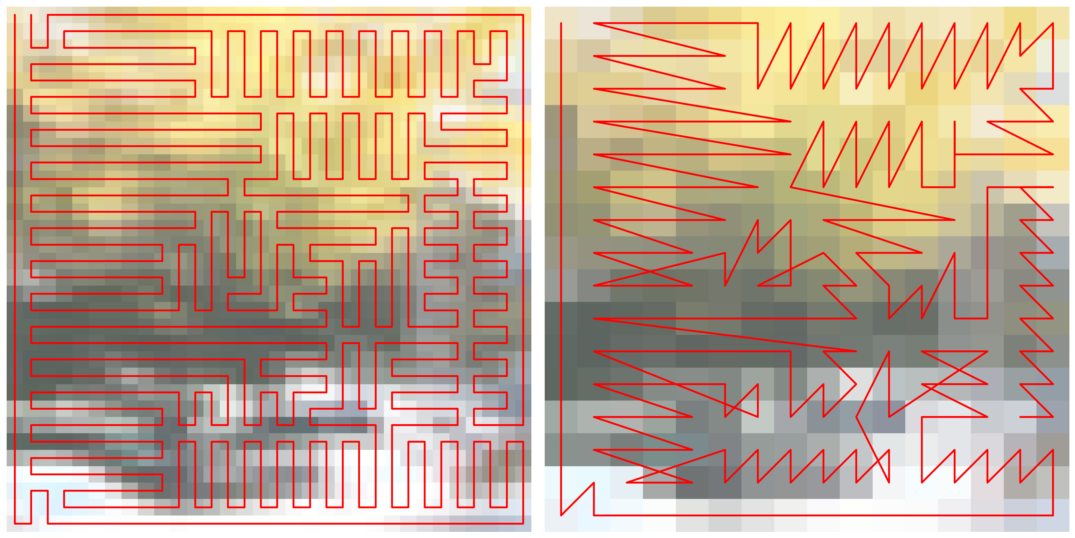}
    \caption{Visual comparison between an SFC curve and its downscaled counterpart. Downscaling the SFC leads to a large number of non-contiguous jumps.}
    \label{fig_traditional_down}
   \vspace{-1mm}
\end{figure}

As discussed in Sec.~\ref{section_method_scaling}, directly downscaling an SFC can disrupt locality and introduce non-contiguous jumps along the path. The traditional downscaling approach maps each fine level $(r, c)$ coordinate along the path to a coarse cell via integer division by downscaling factor $s$. The coarse SFC is then formed by listing coarse blocks in the order they are first entered when traversing through the original fine SFC. The algorithm of traditional SFC downscaling and its outputs are provided in Algorithm.~\ref{alg:sfc_downscaling_traditional} and Fig.~\ref{fig_traditional_down}, which are notably inferior to that in Fig.\ref{fig_teaser}\textcolor{red}{III}.

\begin{algorithm}[tb]
\caption{Traditional Curve Downscaling}
\label{alg:sfc_downscaling_traditional}
\centering
\begin{algorithmic}[1]
\STATE \textbf{Input:} Fine-level SFC $\mathcal{S}$, $s =(r,c)$ where $r$ is row and $c$ is column on an $H \times W$ grid , downscale factor $d$
\STATE \textbf{Output:} Coarse SFC $\mathcal{S}_d$, where ${S}_c=(R_d,C_d)$
\STATE {// map fine coordinates to coarse cells in SFC order, $N = H * W$}
\FOR{$i \gets 0$ \textbf{to} $N-1$}
    \STATE $(r_i,c_i) \gets \mathcal{S}[i]$
    \STATE $R_i \gets  r_i // d$
    \STATE $C_i \gets c_i // d $
    \STATE $S_c[i] \gets (R_i,C_i)$
\ENDFOR

\STATE $V \leftarrow \emptyset$ {// visited set}
\STATE $F \leftarrow \{\}$ {// first fine index where coarse cell appears}
\STATE $S_d \leftarrow [\,]$ {// list of kept indices}
\FOR{$i \gets 0$ \textbf{to} $N-1$}
    \STATE $(R_i,C_i) \gets S_c[i]$
    \IF{$(R_i,C_i) \notin V$}
        \STATE $V \gets V \cup \{(R_i,C_i)\}$
        \STATE $F[(R_i,C_i)] \gets i$ // record first-visit position
        \STATE Add $(R_i,C_i)$ to $\mathcal{S}_d$
    \ENDIF
\ENDFOR

\STATE \textbf{return} $\mathcal{S}_d$
\end{algorithmic}
\end{algorithm}

\input{sec/2_related}

\section{More Visual Companions on Benchmark Datasets}

\begin{figure*}[t]
    \setlength{\abovecaptionskip}{0.05in}
    \centering\setlength{\baselineskip}{0mm}
    \vspace{0mm}
    \includegraphics[width=1\linewidth]{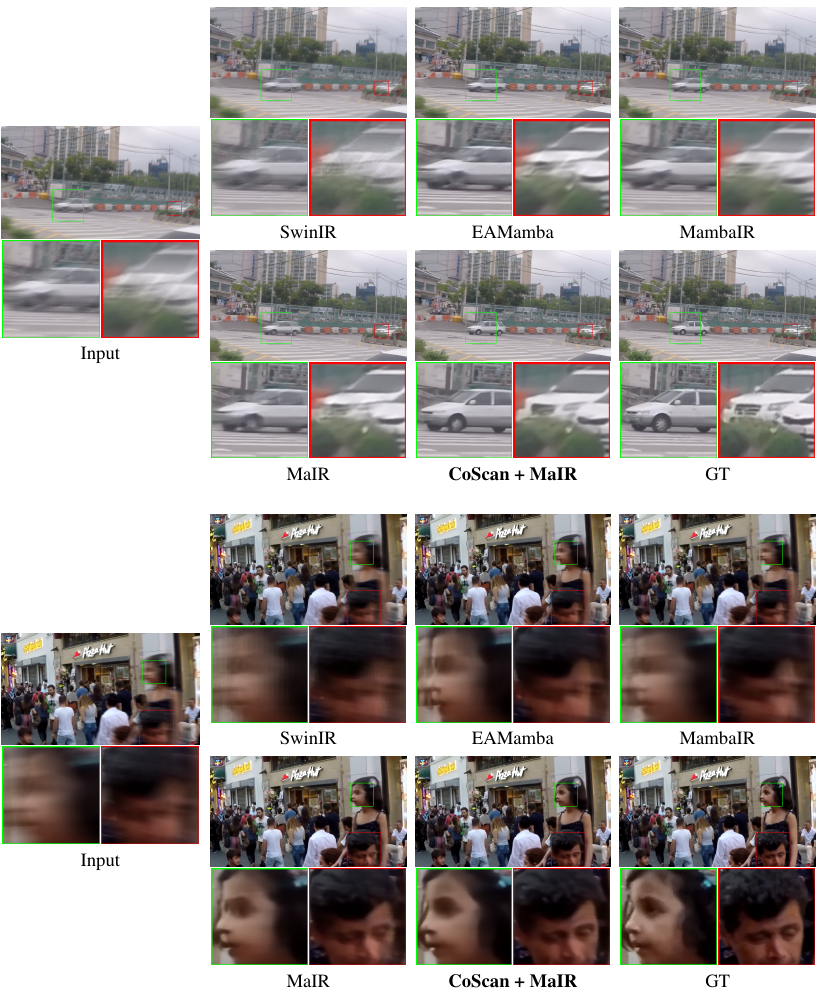}
    \caption{Additional visualizations on GoPro dataset (deblurring task). Compared to other baselines, our method faithfully restores detailed object and facial features while avoiding ghosting artifacts. In contrast, ghosting in baselines obscure identity-relevant features such as hair and eyes in Rows 3 and 4.}
    \label{fig_vis_gopro_1_appendix}
    \vspace{-2mm}
\end{figure*}

\begin{figure*}[t]
    \setlength{\abovecaptionskip}{0.05in}
    \centering\setlength{\baselineskip}{0mm}
    \vspace{0mm}
    \includegraphics[width=1\linewidth]{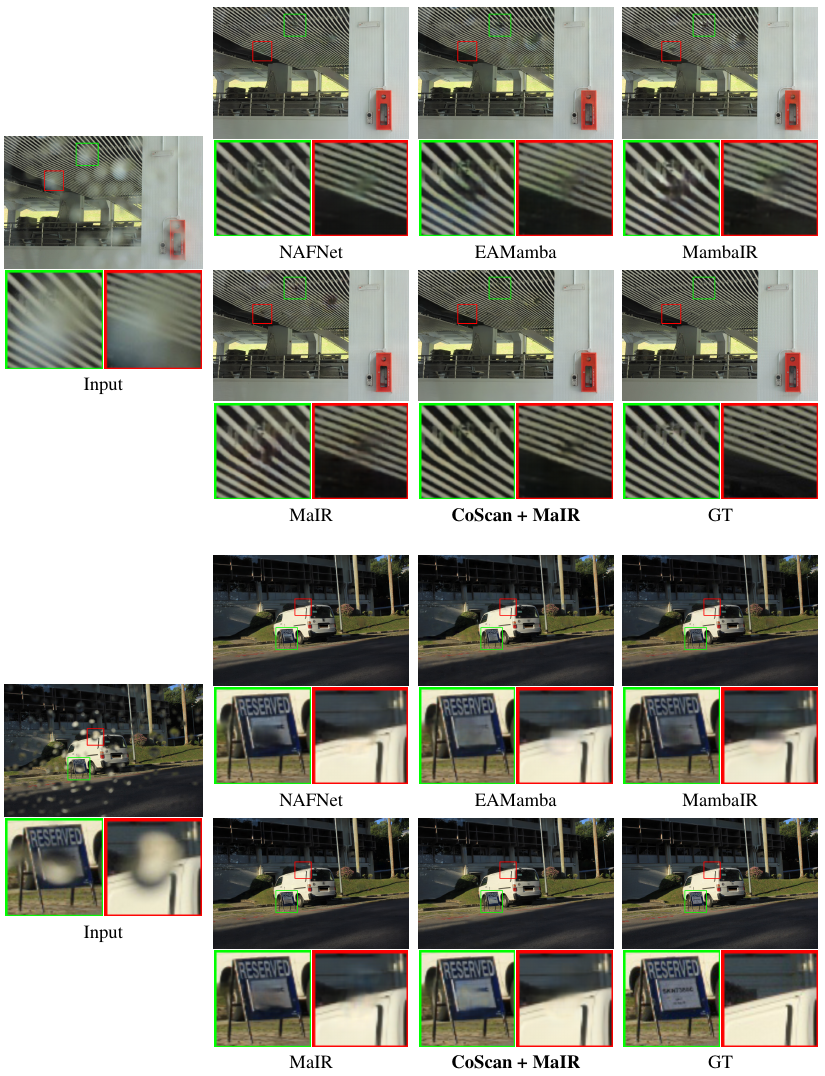}
    \caption{More visualizations on the raindrop removal task. Raindrop occlusions in baseline methods produce smeared and dark artifacts, whereas our method more effectively preserves stronger edges and maintains original scene colors in the highlighted regions.}
    \label{fig_vis_raindrop_appendix}
    \vspace{-2mm}
\end{figure*}

In Figs. \ref{fig_vis_gopro_1_appendix} and
\ref{fig_vis_raindrop_appendix}, we provide extensive visual evidence that demonstrates the efficacy of our CoScan on different datasets, highlighting its superiority over existing image restoration methods.

\section{Limitations and Future Work}
\label{sec_limitation}
Following existing Mamba-based image restoration frameworks~\cite{guo2024mambair, li2025mair}, CoScan currently performs scan generation and inference in a patch-wise manner for high-resolution images. While this design maintains practical efficiency and stable memory usage, it may limit the modeling of extremely long-range dependencies across distant image regions, since each patch is processed independently with its own scan ordering. Future work may explore more globally coordinated scan construction strategies, such as hierarchical coarse-to-fine scan organization across neighboring patches, to enable more effective long-range context propagation for ultra-high-resolution restoration. 

\newpage

%% file: sec/2_related.tex
\section{Related Works}

\subsection{Non-Mamba Methods for Image Restoration}

\noindent \textbf{CNNs:} Early CNN works \citep{mao2016image, zhang2017beyond, tai2017image} introduce encoder-decoder networks and residual learning for image restoration tasks. MPRNet \citep{zamir2021multi} further presents a multi-stage U-Net architecture where cross-stage feature fusion progressively refines high and low-level features. NAFNet \citep{nafnet} removes nonlinear activations with simplified channel attention and gating mechanism, achieving SOTA performance. However, due to their local receptive fields, CNNs remain less naturally suited for modeling long-range dependencies. \textbf{Transformers:} Transformers address this shortcoming using self-attention \citep{dong2024shadowrefiner, dong2025sgllie,dong2024dehazedct}. IPT \citep{chen2021pre} introduces vision transformers with learnable task embeddings for image restoration. To reduce quadratic cost and memory demands of self-attention, Restormer \citep{zamir2022restormer} proposes a channel-wise attention transformer with depth-wise convolutions. Hybrid-attention transformer blocks \citep{chen2023hat, zeng2025all} have demonstrated improved performance through multifaceted attention mechanisms. However, although these methods achieve strong restoration performance through different attention variants, their reliance on QKV operations still introduces substantial computation overhead. \textbf{Generative:} Generative approaches have achieved widespread success in image restoration \citep{gppllie}. Early score-based generative models \citep{song2020score} and denoising diffusion probabilistic models (DDPM) \citep{ho2020denoising} established the foundation for diffusion-based image generation and restoration. Building upon these developments, IR-SDE \citep{luo2023image} introduces a mean-reverting process that more effectively models image degradation. Later works further improve performance by directly diffusing priors or residuals \citep{xia2023diffir}, as well as by adopting prior-guided diffusion models \citep{saharia2022palette}. More recently, VAR- and flow matching-based generative models \cite{varsr, restorevar, PMRF, dong2026zero, dong2026pvrf} have emerged as promising directions for image restoration. However, they typically require iterative multiple steps during inference, leading to high inference latency even with distillation. In this work, we focus on improving sequence modeling through a coherence-aware and causality-aligned scanning strategy tailored for Mamba.

%\noindent \textbf{Mamba-based Methods:} 
\subsection{Vision Mamba for Image Restoration}
Mamba \citep{gu2024mamba} is a recently proposed State Space Model (SSM) with input-dependent gating and selective scanning to achieve efficient long-range context, while its parallel recurrent structure serves as a promising alternative to transformers. Vim \citep{zhu2024vision} and VMamba \citep{liu2024vmamba} introduce bidirectional and cross-scanning to reduce directional bias in high-level vision tasks. Adopting earlier Mamba works, specific image restoration tasks have been a major focus, including but not limited to super-resolution \citep{qiao2024hi, lei2024dvmsr}, dehazing \citep{zheng2024u}, exposure correction \citep{dong2024ecmamba}, and deblurring \citep{kong2025efficient, gao2024learning, liu2025xyscannet}. Pioneering task-adaptive image restoration, MambaIR \citep{guo2024mambair} applies both bidirectional and cross-scanning methods to low-level vision, adding an attention-based refinement module to help guide image restoration. MaIR \citep{li2025mair} further explores scanning methods by employing nested s-shaped scanning paths with a shift-stripe mechanism to ensure better locality and continuity. However, to our knowledge, these methods lack exploration of a truly dynamic, spatially content-adaptive scanning method.

\noindent \subsection{Scanning in State Space Models}
In SSM-based methods, the scanning strategy for flattening 2D images into 1D sequences critically affects feature representation as earlier tokens reduce state decay and adjacent tokens preserve locality. Existing scanning methods can be categorized as fixed or dynamic. Fixed strategies include row-wise scanning in U-Mamba \citep{ma2024u}, bidirectional row and column-wise scanning in Vim and VMamba \citep{zhu2024vision, liu2024vmamba}, nested s-shaped scanning in MaIR \citep{li2025mair}, and fractal-based (Hilbert) scanning in RainMamba \citep{wu2024rainmamba} and FractalMamba \citep{xiao2025boosting}, all of which aim to improve local and global feature representation in 1D sequences. However, fixed scanning often fails to capture complex structural relationships and variations in an image.
To address this issue, MambaIRv2 \citep{guo2025mambairv2} proposes a dynamic scanning method, using semantic guided neighboring (SGN) that reorders tokens based on the image, ensuring semantically similar pixels stay adjacent. DAMamba \citep{li2025damamba} similarly addresses this problem with dynamic adaptive scan (DAS), which route sequences via predicted landmark points. Nonetheless, existing dynamic scanning methods can produce non-contiguous jumps and selection bias, weakening spatial locality and diluting global structure.